\documentclass[pdflatex,sn-nature]{sn-jnl}
\setcitestyle{super,open={},close={}}

\usepackage{graphicx}
\usepackage{multirow}
\usepackage{amsmath,amssymb,amsfonts}
\usepackage{booktabs}
\usepackage{makecell}
\usepackage{silence}
\usepackage[labelformat=simple]{subcaption}

\newlength{\panelw}
\newsavebox{\panelbox}
\newcommand{\panel}[2]{\sbox\panelbox{#2}\makebox[\panelw][l]{\raisebox{\dimexpr\ht\panelbox-0.75em\relax}{\sffamily\bfseries #1}}\usebox\panelbox}

\newcommand{\relltwo}{\ensuremath{\text{rel-}L_2}}
\newcommand{\DEL}{\textsc{del}}
\newcommand{\Uhub}{\ensuremath{U_{\mathrm{hub}}}}
\newcommand{\Gop}{\ensuremath{\mathcal{G}}}

\let\table\tableorg\let\endtable\endtableorg
\makeatletter\AtBeginDocument{\def\toclevel@subparagraph{1}}\makeatother

\begin{document}

\title[Hybrid state-space and attention neural operators]{Hybrid state-space and attention neural operators for nonlinear structural dynamics under environmental loading}

\author[1,3]{\fnm{Anastasios} \sur{Stamou} }
\author[2]{\fnm{Rui} \sur{Teixeira}}
\author*[3]{\fnm{Maria} \sur{Nogal}}\email{M.Nogal@tudelft.nl}
\author*[1]{\fnm{Michalis} \sur{Fragiadakis}}\email{mfrag@mail.ntua.gr}

\affil[1]{\orgname{National Technical University of Athens}, \orgaddress{\street{9 Iroon Polytechniou, Zografou Campus}, \city{Athens}, \postcode{15780}, \country{Greece}}}
\affil[2]{\orgdiv{Department of Civil, Structural and Environmental Engineering}, \orgname{Trinity College Dublin}, \orgaddress{\street{Museum Building, College Green}, \city{Dublin 2}, \postcode{D02 PN40}, \country{Ireland}}}
\affil[3]{\orgdiv{Faculty of Civil Engineering and Geosciences}, \orgname{Delft University of Technology}, \orgaddress{\street{Stevinweg 1}, \city{Delft}, \postcode{2628 CN}, \country{The Netherlands}}}

\abstract{The computational burden of repeated nonlinear time-history analyses limits their application for structural assessment and design.
We therefore propose DeepOJamba, a hybrid neural operator that combines selective state-space recurrence with full attention to predict the structural response history due to dynamic loading.
Physical-parameter conditioning and predicted-state transfer between sequences enable a single trained operator to represent varying system conditions and generate response histories beyond its training sequence length.
DeepOJamba is evaluated on a 15~MW monopile offshore wind turbine under wind and wave loading.
The proposed operator is compared to pure attention, pure state-space, Fourier, and DeepONet operators and is found to give the lowest window error on every channel and the lowest closed-loop fatigue and ultimate-load errors of $9.4\%$ and $6.0\%$, against $14.6\%$ and $8.7\%$ for the next-best pure attention operator examined.
Moreover, on sea states outside the training wave band, the error slightly increases by approximately $0.4$ percentage points.
Pure state-space operators can approach similar low errors only with a much wider convolution kernel tuned to the tower eigenperiod, which increases the training cost by $40\%$.
In all, the results obtained support hybrid neural operators for robust long-duration structural response and demand prediction.}

\keywords{neural operators, state-space models, attention, structural dynamics, fatigue, offshore wind turbines}

\maketitle

\section{Introduction}
\label{sec:intro}

Nonlinear response-history analysis (NLRHA) is the reference method for assessing structures under dynamic and environmental loading.
NLRHA has many applications in engineering, e.g., in probabilistic or optimization-based applications, where it is commonly required to perform many simulations for different structural properties and hazard conditions.
Since every simulation involves thousands of time steps, and a study may require thousands of simulations, the computational cost quickly becomes prohibitive for tackling complex and demanding engineering problems.

Surrogate models reduce the cost of NLRHA by replacing the high-fidelity model with an inexpensive approximation, so that the many evaluations needed to estimate response statistics become affordable.
Most surrogates are trained to predict a small set of scalar quantities of interest (QoIs), such as peak demands or damage measures.
In wind-turbine load assessment, Gaussian-process regression (Kriging), polynomial chaos expansions, and related methods are well established \citep{slot_surrogate_2020,dimitrov_wind_2018,teixeira_stress-cycle_2019,wilkie_gaussian_2021}.
With scalar-output surrogates, however, the QoIs have to be chosen before training, and any quantity not selected in advance requires retraining the model.
A surrogate that predicts the full response history avoids this limitation.

Operator learning provides a framework for predicting full response histories.
It treats the surrogate as a mapping between function spaces, from loading histories, physical parameters, and initial conditions of the structure to the resulting response history \citep{lu_deeponet_2021,li_fourier_2021,kovachki_neural_2024,kontolati_learning_2024}.
QoIs such as peak values, response spectra, cycle ranges, or fatigue damage are then extracted from the predicted histories by standard post-processing, so a new QoI requires no retraining.
The predicted waveforms are also easier to interpret and assess: inspecting them helps to distinguish amplitude, phase, and hysteretic errors, which a scalar prediction can conceal.
Finally, if the operator is trained over a range of structural properties and loading conditions, a single surrogate can represent a whole family of structures, so the training cost is shared across related simulation campaigns.

Several neural-operator architectures can learn the mapping from loading to response histories.
They differ mainly in how they represent dependence on past loading.
The deep operator network (DeepONet) and the Fourier neural operator (FNO) are two of the most established formulations.
DeepONet combines an encoder of the loading history with a learned temporal basis \citep{lu_deeponet_2021}, whereas FNO mixes the sampled histories through convolutions in the Fourier domain \citep{li_fourier_2021}.
Both treat the history globally rather than step by step.
Sequence models offer two alternative ways to represent temporal dependence.
Attention connects all time steps within a window directly \citep{vaswani_attention_2023} and has been used to learn the dynamics of physical systems \citep{casert_learning_2024,gao_generative_2024}.
Attention cost grows quadratically with sequence length, which has motivated linearized variants such as the Galerkin Transformer and the General Neural Operator Transformer \citep{cao_choose_2021,hao_gnot_2023}.
Recurrent and state-space models instead carry information forward through a latent state, as in sequence-to-sequence forecasting of seismic wavefields \citep{lyu_rapid_2025}.
Among state-space models, Mamba uses input-dependent state updates and a scan whose cost is linear in sequence length \citep{gu_mamba_2024}.
Mamba-based operators have been applied with success to partial differential equations and dynamical systems \citep{hu_deepomamba_2025,zheng_alias-free_2024,cheng_mamba_2026,hu_state-space_2026}.
Hybrid models such as Jamba combine the two mechanisms by alternating state-space and attention layers, and in language modeling they have been reported to outperform both pure architectures \citep{lieber_jamba_2024,waleffe_empirical_2024}.
To the authors' knowledge, such hybrids have not yet been used as neural operators.

In structural engineering, recurrent networks and neural operators have been applied to nonlinear seismic response prediction \citep{zhang_deep_2019,goswami_neural_2025,hu_struct-wno_2025}.
These studies evaluate the predicted waveforms, but they do not move on to the assessment of design demand parameters.
For example, Goswami et al.~\citep{goswami_neural_2025} assess structural-response operators by mean squared and relative $L_2$ (\relltwo{}) errors.
These metrics do not directly measure the accuracy of peak or fatigue demands.
Moreover, when such pointwise errors are used as training losses, they tend to favor smoothed predictions if the target is uncertain \citep{mathieu_deep_2016,oommen_learning_2026}, while smoothing reduces exactly the peaks and cycle ranges that govern these demands.
For experimentally tested bridge bearings and piers, Yazdanpanah et al.~\citep{yazdanpanah_mastering_2024} embedded attention and Mamba layers in a model that combines a convolutional neural network (CNN) with a bidirectional long short-term memory network, with ground motion, measured actuator-force histories, and structural descriptors as inputs.
Their evaluation includes peak displacements and hysteretic energy, but not rainflow-based damage-equivalent loads (\DEL{}s) or ultimate limit state (ULS) load demands.
In wind energy, surrogate applications include an autoregressive emulator of blade-root bending moments driven by the turbulent wind field and evaluated by waveform error and peak moment~\citep{schar_emulating_2024}, virtual sensors that estimate loads from measured signals~\citep{dimitrov_virtual_2022}, and real-time surrogates of the operating mode and tower oscillations coupled with the turbine controller~\citep{witter_creating_2025}.

In contrast to measurement-driven virtual sensors, the present work develops surrogates driven by the prescribed loading.
The proposed models are conditioned on physical descriptors such as the wave direction and can start from any physically plausible initial state.
A single model therefore covers a range of conditions and can be chained to simulate longer durations.
We evaluate the models on three levels: ($i$) the predicted histories, ($ii$) the peak and fatigue demands derived from them, and ($iii$) the error accumulated across chained windows.

Models that generalize across a family of structures link structural operator learning to broader efforts towards scientific \textit{foundation models}.
Foundation models are large models that are trained across parameter ranges, governing equations, and physical domains, so that they cover a class of problems \citep{subramanian_towards_2023,mccabe_multiple_2024,hao_dpot_2024,herde_poseidon_2024}.
A foundation model for structural dynamics would need to capture how physical properties, operating conditions, and the loads shape the structural response.
It would also need to generalize across initial states and carry the dynamic state from one prediction window to the next, so that long response histories can be generated on standard hardware.
For an Offshore Wind Turbine (OWT), such a model would span the parked, operating, and damaged states, which differ in aerodynamic damping, excitation, and stiffness.
As a step in this direction, this work examines these capabilities for a single OWT across a wide range of environmental and power-production conditions.

This paper introduces DeepOJamba, a Jamba-inspired \citep{lieber_jamba_2024} neural operator that combines Mamba blocks with full attention layers.
The two layer types are complementary.
Mamba mixes neighboring samples through a short causal convolution, four samples wide by default \citep{gu_mamba_2024}, and carries longer-range information on the loading path through its recurrent state.
Attention, by contrast, connects every pair of time steps within the input window directly.
In the combined architecture, attention layers create direct global connections that are propagated by the subsequent state-space blocks.
We show how this additional mixing retains most of the computational and accuracy benefits reported for Mamba operators \citep{hu_deepomamba_2025,hu_state-space_2026}, while improving accuracy for lightly damped, multi-mode responses.
Such responses decay slowly over many resonant cycles, which a compressed recurrent state may struggle to retain.
The model is designed to generalize across environmental conditions, wave directions, and controller operating regimes.
To this end, physical descriptors condition the backbone through adaptive layer normalization with zero initialization (AdaLN-Zero) \citep{peebles_scalable_2023}.
In addition, each prediction window receives an initial structural state, so that the model can start from nonzero conditions. Long response histories are then generated by autoregressive rollout, with the final state of one window initializing the next.

The study has four main contributions.
First, it develops a parameter-conditioned hybrid operator with state transfer between windows for long response histories, that is, histories longer than the sequences used in training.
Second, it compares Fourier, branch--trunk, attention, state-space, and hybrid operators with comparable numbers of trainable parameters on the OWT case study, including held-out sea states outside the training wave band to test generalization to unseen loading.
Third, it evaluates trajectory errors together with peak and cyclic demands, and uses handoff sensitivity studies and a state-perturbation test to separate errors within a window from their accumulation during rollout.
Fourth, since the default Mamba convolution spans only a small fraction of a structural eigenperiod, the paper examines whether widening the convolution to one tower eigenperiod improves the pure state-space and the hybrid operators, and at what training cost.

The results show that, at the default convolution width, the hybrid operator is the most accurate across all examined metrics.
Pure Mamba and attention are less accurate both for single-window trajectories and for engineering metrics post-processed from longer closed-loop predictions.
Pure Mamba also amplifies errors in the initial states it receives between windows.
The accuracy of the hybrid operator is insensitive to the convolution width, so it does not need width tuning.
Finally, a one-eigenperiod convolution kernel suppresses most of the error amplification in pure Mamba and brings it close to the hybrid operator's fatigue and ultimate-load accuracy, but increases the training cost by $40\%$.

\section{Results}
\label{sec:results}

\subsection{Case study and model description}
The case study is the International Energy Agency (IEA) 15~MW reference OWT on a monopile~\citep{gaertner_definition_2020}, subjected to turbulent wind and irregular waves (Fig.~\ref{fig:overview:case}).
It is a coupled aero-hydro-servo-elastic system in which aerodynamic and hydrodynamic loads interact dynamically with the oscillating system.
The OWT's lightly damped modes make this a demanding benchmark for long response histories, since the response history is affected by the earlier time steps, while small prediction errors often persist across successive windows and do not fade away.
The three-bladed, upwind, direct-drive turbine has variable rotor speed ($5.0$--$7.56$~rpm) and collective blade-pitch control (dimensions in Fig.~\ref{fig:overview:case}).
The blade-pitch/torque schedule reaches rated operation near $10.5$~m/s.

The proposed surrogate model predicts the time-histories of eight design load channels (Supplementary Table~1).
Fatigue and ULS demands can be post-processed in order to characterize the response.
The monopile is clamped to the seabed.
The predicted quantities are the fore--aft and the side--side bending moments at the tower base, and also
the ``fore--aft'' and ``side--side'' shear forces and overturning moments at the mudline.
We also monitor, at the root of one blade, the flapwise and edgewise bending moments.
Fore--aft (FA) denotes the mean wind direction and side--side (SS) the horizontal direction normal to it.
The FA loads are driven directly by the rotor thrust, and the operating rotor damps them aerodynamically.
The SS loads receive little aerodynamic damping and, unless the waves are misaligned with the wind, they are only weakly forced.
They therefore respond mainly at the lightly damped first SS tower mode.
The flapwise moment acts out of the rotor plane and follows the aerodynamic load on the blade.
The edgewise moment acts in the rotor plane and is dominated by the blade weight, whose moment reverses once per revolution.
The surrogate also predicts thirteen state channels and the rotor phase, which start each subsequent prediction window (controller, platform, tower, and nacelle states; Supplementary Table~1).

The OWT is simulated with OpenFAST~v4.2.1~\citep{openfast} and the ROSCO~v2.10.4 controller~\citep{abbas_reference_2022} (Turbine model and simulation settings in Supplementary Methods).
Each simulation lasts $600$~s, with outputs written every $\Delta t = 10$~ms, giving $60{,}001$ samples.
The response is lightly damped and multi-modal: tower FA and SS bending near
$0.18$~Hz, once- and three-per-revolution rotor harmonics, wave-frequency forcing near $0.1$~Hz, and blade modes coexist within each record.
The operational SS damping is $0.46\%$ of critical (decay time $199$~s; Supplementary Methods - Turbine model and simulation settings), against $9.4\%$ in the FA direction.

We choose to showcase the proposed model capabilities on the OWT because it is both costly to simulate and has a complex response for the model to learn.
Wind-turbine design requires ultimate and fatigue limit-state checks over a large matrix of wind, wave, and operating conditions, each simulated with several turbulence and wave seeds \citep{iec61400-1,iec61400-3}.
Reliability and sensitivity studies add to this cost, reaching $98{,}400$ HAWC2 \citep{mozafari_sensitivity_2024} and $324{,}000$ OpenFAST \citep{wiley_sensitivity_2023} simulations.
Fatigue is assessed by rainflow counting and Palmgren--Miner summation over the complete load history \citep{matsuichi_fatigue_1968,downing_simple_1982,miner_cumulative_2021}.
A surrogate of the full history therefore leaves the hotspot, W\"ohler exponent and counting method open, whereas surrogates of \DEL{}s fix them at training time.
Finally, the SS decay time ($199$~s) is a third of the record length, which makes the case a stringent test of long-range temporal memory.

\subsection{Environmental loading and dataset}
\label{sec:res:data}

The response of each simulation is driven by a turbulent wind field and an irregular sea state with a prescribed wind--wave misalignment angle.
The wind loading is a full-field turbulent inflow generated with TurbSim software~\citep{jonkman_turbsim_2009} on a $21\times21$ grid spanning $252$~m.
The wave loading is an irregular, long-crested sea with a single mean heading and is described by a JONSWAP (Joint North Sea Wave Project) spectrum~\citep{hasselmann_measurements_1973}.
The spectrum is parameterized by the significant wave height $H_s$ and the peak period $T_p$.
The wind--wave misalignment angle is introduced by sampling the wave heading relative to the fixed nacelle direction.
Varying the misalignment angle changes the direction of the wave excitation towards the lightly damped SS direction and therefore tests the operator's ability to adjust the response to the loading direction.
Structural parameter variability was not considered.

In total, $2960$ simulations were performed: $1803$ for training, $257$ for validation, $450$ for in-distribution testing, and $450$ for out-of-distribution testing on held-out wave bands (Section~\ref{sec:res:ood}).
Four environmental parameters are sampled uniformly using Latin Hypercube Sampling~\citep{mckay_comparison_1979}: the hub-height wind speed $\Uhub\in[3,25]$~m/s, which covers the full power-producing range of the turbine, the significant wave height $H_s\in[0.5,9]$~m, the peak period $T_p\in[3,18]$~s, and the wind--wave misalignment angle $\theta_{\mathrm{mis}}\in[-90^\circ,90^\circ]$.
Except for the heading-interpolation anchors, each simulation uses its own TurbSim and wave seeds, so every metocean point is a single stochastic realization.
The training, validation, and in-distribution test sets are confined to a moderate-steepness $T_p$--$H_s$ band.
The in-distribution test set contains $250$ independent Latin-hypercube cases, $50$ near-aligned cases drawn the same way but with $|\theta_{\mathrm{mis}}|<5^\circ$, and $150$ heading-interpolation cases, each of which differs from two training-pool simulations only in wave heading.
Supplementary Fig.~1 shows the resulting coverage, and the Supplementary Methods - Environmental loading and dataset paragraph gives the remaining details of the turbine model, the loading, and the dataset.

\subsection{Surrogate architecture}
\label{sec:res:setup}

\begin{figure}[htbp]
\centering
\begin{subfigure}{\linewidth}\centering
  \includegraphics{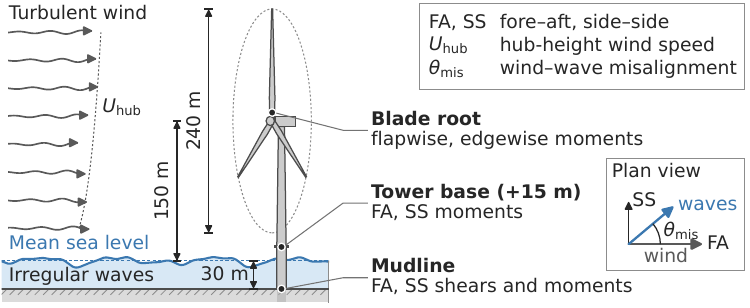}\caption{}\label{fig:overview:case}
\end{subfigure}\\
\begin{subfigure}{\linewidth}\centering
  \includegraphics{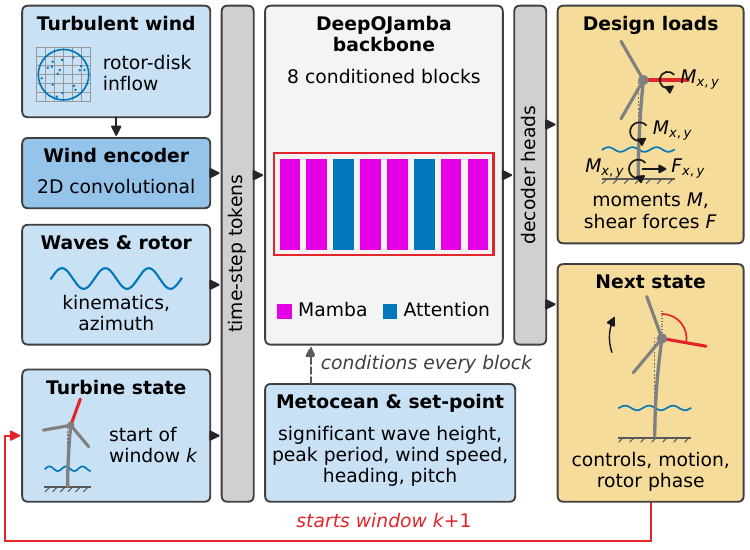}\caption{}\label{fig:overview:pipe}
\end{subfigure}
\caption{OWT and surrogate architecture:
\textbf{a}, IEA~15~MW monopile OWT,
\textbf{b}, architecture of DeepOJamba neural-operator surrogate.
}
\label{fig:overview}
\end{figure}

Figure~\ref{fig:overview}b summarizes the surrogate architecture.
The surrogate takes two inputs: a time-dependent token stream and a fixed simulation descriptor.
The token stream combines three parts at each time step.
First, a small two-dimensional CNN compresses the turbulent wind field inside the rotor disk, sampled at $293$ grid points with three velocity components each, to $63$ total features.
Second, a $42$-channel forcing vector adds wave, wind-grid, and rotor quantities (Supplementary Methods - Surrogate inputs).
Third, the $13$-channel state at the start of the window is repeated at every time step, so that each token carries the initial conditions, giving $118$ channels per token in total.
The simulation descriptor holds six values: the normalized significant wave height $H_s$, peak period $T_p$, and hub-height mean wind speed $\Uhub$, the wave heading encoded as $\sin/\cos$, and the steady-state pitch set-point.
Through AdaLN-Zero, these descriptors condition every block of the backbone on the prescribed regime.
The attention (DeepOT), state-space (DeepOMamba), and proposed hybrid operators share the same inputs, conditioning, and output heads, and differ only in the temporal backbone (Fig.~\ref{fig:overview}).
Because the dynamics of the FA and the SS response differ, a three-head decoder predicts the FA, SS, and remaining channels of the $21$-channel output separately (Supplementary Table~1).
A small auxiliary head returns the rotor azimuth as a $\sin/\cos$ pair.

DeepOJamba uses six Mamba (M) and two full-attention (A) blocks arranged in a \texttt{MMAMMAMM} pattern.
Each attention block spans the complete $4096$-step ($40.96$~s) window.
Each Mamba block first mixes neighboring past time steps through a causal convolution of width $d_{\mathrm{conv}}$, set to the Mamba default $d_{\mathrm{conv}}=4$ unless stated otherwise.
In the kernel-width study, a subscript denotes the convolution width, e.g., DeepOMamba$_{565}$ uses a $565$-step kernel spanning one $5.65$~s tower eigenperiod.

The operators used for comparison differ from DeepOJamba mainly in their temporal backbone.
To isolate the contribution of the backbone, DeepOJamba is first compared with two sequence operators of similar size ($5.9$--$6.2$\,M parameters): DeepOT, which uses nine full-attention blocks, and DeepOMamba, which uses only Mamba blocks.
FNO and DeepONet are also included as spectral and branch--trunk baselines, with the same inputs, wind encoder, and training recipe.
FNO uses the same output heads as the sequence operators, whereas DeepONet returns all channels through its separable branch--trunk expansion.

Long response histories are predicted as a rollout: the record is divided into consecutive segments (``windows''), which the operator predicts one at a time.
In a \textit{closed-loop} rollout, each window starts from the state the operator itself predicted in the previous window, so prediction errors can propagate from one window to its next.
In an \textit{open-loop} evaluation, by contrast, every window starts from the reference state, which measures single-window accuracy but hides error accumulation.
In practice, no reference simulation is available when a surrogate is used, and thus closed-loop rollout must be followed.

Unless stated otherwise, each $600$~s record is predicted by a closed-loop rollout of $15$ consecutive $4096$-step windows.
By default, the windows do not overlap: each window runs to its last step, and the state predicted there becomes the initial state of the next window (end-to-end handoff; Section~\ref{sec:method:chain}).
Handing off at an earlier step makes consecutive windows overlap, which can affect both accuracy and inference cost. Alternative handoff locations are examined in Supplementary Table~2.
Following common practice in OWT load assessment, all metrics exclude the initial $100$~s start-up transient.
Supplementary Fig.~2 shows that the operator also reproduces this transient in a rollout started at $t=0$, so it can be initialized without first running a numerical model.

All models are evaluated on three measures: the error of the predicted response history within a window, fatigue error \DEL{}, and the ultimate load derived from the closed-loop histories.
The training hyperparameters and the model selection scheme are given in Supplementary Table~3.
Each operator configuration is trained with three seeds (Supplementary Methods - Training windows and seeds).
All training runs share the same data split and are evaluated on all $450$ test simulations, with the mean $\pm$ one standard deviation (s.d.) describing the variation across the trained models.

\subsection{Response prediction}

\subsubsection{Windowed prediction}
\label{sec:res:summary}
\label{sec:res:window}

\begin{figure}[htbp]
\centering
\includegraphics{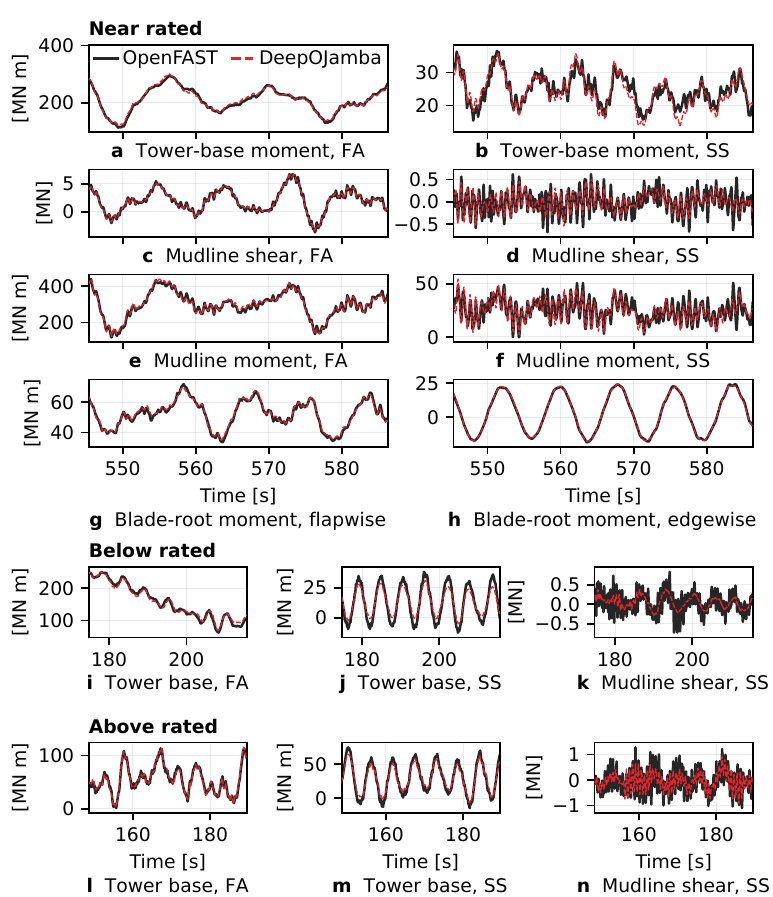}
\caption{\textbf{Windowed response in three operating regimes.} \textbf{a}--\textbf{h}, The eight load channels near rated wind speed ($10.6$~m/s), from OpenFAST and DeepOJamba, one prediction window started from the ground truth initial state. \textbf{i}--\textbf{n}, Tower-base moments and mudline side--side shear below ($8.1$~m/s) and above ($21.0$~m/s) rated wind speed. All windows have near-aligned waves and an error close to the test-set median.}
\label{fig:owt:win}
\end{figure}

Table~\ref{tab:owt:val} and Figure~\ref{fig:owt:win} compare single prediction windows of DeepOJamba with the OpenFAST reference.
Each window starts from the reference initial state, so the figure shows the accuracy before window chaining.
Three wind speed operating conditions are shown: near-rated, below-rated and above-rated, corresponding to wind speeds of $10.6$, $8.1$ and $21.0$~m/s, respectively.
In near-rated wind speed (Fig. \ref{fig:owt:win}a--\ref{fig:owt:win}h), the controller switches between torque and pitch control; thus, the OWT behavior is most complex.
The windows shown have errors close to the test-set median, so they represent typical predictions rather than best cases.
The predicted histories follow OpenFAST closely in all three operating wind speed conditions, and the main design loads are reproduced with small error.

\begin{table}[htbp]
\centering
\caption{\textbf{In-distribution window error.} Per-window \relltwo{} over the $450$ test simulations. Mean $\pm$ s.d. over three training seeds.}
\label{tab:owt:val}
\small\setlength{\tabcolsep}{3.5pt}
\setlength{\aboverulesep}{0pt}\setlength{\belowrulesep}{0pt}\renewcommand{\arraystretch}{1.15}
\begin{tabular*}{\linewidth}{@{\extracolsep{\fill}}l cccc|c@{}}
\toprule
Channel & DeepONet & FNO & DeepOT & DeepOMamba & DeepOJamba\\
\midrule
\multicolumn{5}{@{}l|}{\emph{Fore--aft and blade channels}} &\\
Tower-base moment & $0.142\,{\pm.001}$ & $0.095\,{\pm.002}$ & $0.072\,{\pm.000}$ & $0.069\,{\pm.002}$ & $\mathbf{0.061\,{\pm.001}}$\\
Mudline shear & $0.303\,{\pm.001}$ & $0.233\,{\pm.001}$ & $0.170\,{\pm.001}$ & $0.171\,{\pm.002}$ & $\mathbf{0.154\,{\pm.001}}$\\
Mudline moment & $0.165\,{\pm.000}$ & $0.127\,{\pm.001}$ & $0.087\,{\pm.000}$ & $0.087\,{\pm.002}$ & $\mathbf{0.074\,{\pm.001}}$\\
Blade-root edgewise & $0.190\,{\pm.002}$ & $0.141\,{\pm.001}$ & $0.113\,{\pm.000}$ & $0.109\,{\pm.003}$ & $\mathbf{0.101\,{\pm.000}}$\\
Blade-root flapwise & $0.165\,{\pm.000}$ & $0.095\,{\pm.002}$ & $0.062\,{\pm.001}$ & $0.066\,{\pm.001}$ & $\mathbf{0.048\,{\pm.001}}$\\
\multicolumn{5}{@{}l|}{\emph{Side--side channels}} &\\
Tower-base moment & $0.177\,{\pm.000}$ & $0.138\,{\pm.003}$ & $0.116\,{\pm.001}$ & $0.147\,{\pm.014}$ & $\mathbf{0.102\,{\pm.002}}$\\
Mudline shear & $0.550\,{\pm.003}$ & $0.460\,{\pm.003}$ & $0.433\,{\pm.000}$ & $0.391\,{\pm.004}$ & $\mathbf{0.349\,{\pm.006}}$\\
Mudline moment & $0.310\,{\pm.000}$ & $0.279\,{\pm.002}$ & $0.255\,{\pm.000}$ & $0.245\,{\pm.010}$ & $\mathbf{0.197\,{\pm.005}}$\\
\midrule
Mean (8 channels) & $0.250\,{\pm.000}$ & $0.196\,{\pm.002}$ & $0.164\,{\pm.001}$ & $0.161\,{\pm.004}$ & $\mathbf{0.136\,{\pm.002}}$\\
\bottomrule
\end{tabular*}

\end{table}

The FA and blade loads are the most accurately predicted channels.
The FA tower-base and mudline moments have window errors of only $0.061$ and $0.074$ (Table~\ref{tab:owt:val}).
The flapwise blade-root moment has the lowest error of all loads, equal to $0.048$.
The SS loads are more difficult to reproduce.
Even in this more demanding case of the SS loads, DeepOJamba predicts the tower-base SS moment well (Fig.~\ref{fig:owt:win}b; error $0.102$), although its $199$~s decay time spans about five windows and its amplitude and phase depend strongly on the state at the start of the window.
The SS mudline moment and the SS mudline shear are also predicted well (Fig.~\ref{fig:owt:win}d).
Under near-aligned waves, this shear has a small amplitude and mostly high-frequency content; DeepOJamba follows its slower variations but smooths the high-frequency peaks.

Below and above rated wind speed (Fig.~\ref{fig:owt:win}i--\ref{fig:owt:win}n), the tower-base moments are again predicted closely, and the SS mudline shear is again under-predicted.
Overall, the model appears to have learned the different dynamics that define the response in all three power-production regimes.
For the structural state variables used to initialize subsequent windows, agreement with OpenFAST is close for the controller and FA channels, but weaker for the SS tower-top motion and platform surge velocity (Supplementary Fig.~3a--f).

As shown in Table~\ref{tab:owt:val}, over all $450$ test simulations, DeepOJamba has the lowest \relltwo{} window error on every load channel, with a mean of $0.136$ against $0.161$ for DeepOMamba and $0.164$ for DeepOT, reductions of approximately $16\%$ and $17\%$, respectively.
The largest improvements occur in the tower-base SS and blade-root flapwise moments.
FNO and DeepONet have higher mean errors of $0.196$ and $0.250$, respectively, with DeepONet performing worst on every channel.

\subsubsection{Closed-loop prediction}
\label{sec:res:eng}

Engineering accuracy is examined by deriving fatigue \DEL{}s and ultimate (absolute peak) load values from the full $600$~s closed-loop response histories.
Table~\ref{tab:owt:eng} reports the mean of the absolute error of the predicted-to-reference demand ratio $r$, obtained as the mean $|r-1|$ of the $450$ test simulations.

According to Table~\ref{tab:owt:eng}, the trajectory accuracy carries over to the quantities of structural engineering interest.
DeepOJamba leads on all three metrics examined, having the lowest mean fatigue and ultimate-load errors across all eight channels.
For the FA and blade-root moments, DeepOJamba errors are between $2.4\%$ and $6.5\%$.
DeepOMamba has the largest errors among the sequence operators, at $25.7\%$ and $17.4\%$, respectively.
On the three-seed per channel means, DeepOJamba leads on seven of eight fatigue channels and seven of eight ULS channels, including the blade-root flapwise and tower-base SS moments.
DeepOJamba's superior fatigue performance is consistent regardless of the fatigue slopes (Supplementary Methods - Fatigue-slope sensitivity).
However, over complete records, the trajectory \relltwo{} error of DeepOJamba roughly doubles, from $0.136$ per window to $0.271$.
Most of this increase comes from accumulated phase error that comes from small prediction error on the rotor's azimuth (Section~\ref{sec:res:chain}).

\begin{table}[htbp]
\centering
\caption{\textbf{Closed-loop fatigue and ultimate-load errors.} Mean per-simulation error $|r-1|$ in percent over the $450$ test simulations (end-to-end handoff), with $r$ the predicted-to-reference demand ratio. Mean $\pm$ s.d. over three training seeds. Bold: lowest per row.}
\label{tab:owt:eng}
\setlength{\tabcolsep}{3.5pt}
\setlength{\aboverulesep}{0pt}\setlength{\belowrulesep}{0pt}\renewcommand{\arraystretch}{1.15}
\begin{tabular*}{\linewidth}{@{\extracolsep{\fill}}l cccc|c@{}}
\toprule
Channel & DeepONet & FNO & DeepOT & DeepOMamba & DeepOJamba\\
\midrule
\multicolumn{5}{@{}l|}{\textbf{a}\enspace Fatigue \DEL{} error [\%]} &\\
\multicolumn{5}{@{}l|}{\emph{Fore--aft and blade channels}} &\\
Tower-base moment & $25.0\,{\pm0.2}$ & $6.7\,{\pm0.3}$ & $5.9\,{\pm0.3}$ & $4.6\,{\pm1.9}$ & $\mathbf{3.2\,{\pm0.2}}$\\
Mudline shear & $42.9\,{\pm0.0}$ & $30.8\,{\pm1.7}$ & $18.5\,{\pm0.1}$ & $15.7\,{\pm0.9}$ & $\mathbf{14.8\,{\pm0.5}}$\\
Mudline moment & $37.0\,{\pm0.1}$ & $22.2\,{\pm0.7}$ & $11.0\,{\pm0.3}$ & $\mathbf{6.5\,{\pm1.0}}$ & $\mathbf{6.5\,{\pm0.7}}$\\
Blade-root edgewise & $8.5\,{\pm0.1}$ & $4.8\,{\pm0.5}$ & $4.8\,{\pm0.0}$ & $3.6\,{\pm0.6}$ & $\mathbf{3.5\,{\pm0.0}}$\\
Blade-root flapwise & $27.8\,{\pm0.2}$ & $9.2\,{\pm1.0}$ & $7.0\,{\pm0.1}$ & $5.3\,{\pm0.3}$ & $\mathbf{4.0\,{\pm0.1}}$\\
\multicolumn{5}{@{}l|}{\emph{Side--side channels}} &\\
Tower-base moment & $23.6\,{\pm0.5}$ & $34.4\,{\pm2.1}$ & $13.9\,{\pm0.9}$ & $93.7\,{\pm62.5}$ & $\mathbf{11.3\,{\pm0.6}}$\\
Mudline shear & $43.0\,{\pm0.1}$ & $37.0\,{\pm0.3}$ & $31.9\,{\pm0.0}$ & $\mathbf{16.9\,{\pm0.7}}$ & $19.7\,{\pm0.9}$\\
Mudline moment & $35.4\,{\pm0.6}$ & $42.4\,{\pm1.9}$ & $23.8\,{\pm0.6}$ & $58.9\,{\pm42.5}$ & $\mathbf{12.0\,{\pm1.1}}$\\
\cmidrule(l){1-6}
Mean (8 channels) & $30.4\,{\pm0.1}$ & $23.4\,{\pm0.7}$ & $14.6\,{\pm0.2}$ & $25.7\,{\pm13.6}$ & $\mathbf{9.4\,{\pm0.4}}$\\
\midrule[\heavyrulewidth]
\multicolumn{5}{@{}l|}{\textbf{b}\enspace Ultimate $|$peak$|$ error [\%]} &\\
\multicolumn{5}{@{}l|}{\emph{Fore--aft and blade channels}} &\\
Tower-base moment & $8.7\,{\pm0.3}$ & $5.2\,{\pm0.4}$ & $3.3\,{\pm0.1}$ & $2.9\,{\pm0.2}$ & $\mathbf{2.6\,{\pm0.1}}$\\
Mudline shear & $21.0\,{\pm0.3}$ & $13.8\,{\pm0.9}$ & $7.8\,{\pm0.0}$ & $6.8\,{\pm0.3}$ & $\mathbf{6.1\,{\pm0.1}}$\\
Mudline moment & $12.0\,{\pm0.2}$ & $7.8\,{\pm0.2}$ & $4.1\,{\pm0.0}$ & $3.3\,{\pm0.3}$ & $\mathbf{3.2\,{\pm0.0}}$\\
Blade-root edgewise & $8.1\,{\pm0.1}$ & $7.6\,{\pm0.4}$ & $6.4\,{\pm0.1}$ & $5.3\,{\pm0.9}$ & $\mathbf{4.9\,{\pm0.0}}$\\
Blade-root flapwise & $12.5\,{\pm0.3}$ & $5.1\,{\pm0.1}$ & $3.3\,{\pm0.1}$ & $2.5\,{\pm0.2}$ & $\mathbf{2.4\,{\pm0.1}}$\\
\multicolumn{5}{@{}l|}{\emph{Side--side channels}} &\\
Tower-base moment & $16.0\,{\pm0.5}$ & $21.0\,{\pm1.0}$ & $8.6\,{\pm0.6}$ & $61.7\,{\pm40.7}$ & $\mathbf{7.5\,{\pm0.4}}$\\
Mudline shear & $30.7\,{\pm0.1}$ & $26.2\,{\pm0.5}$ & $21.7\,{\pm0.1}$ & $\mathbf{12.2\,{\pm2.7}}$ & $12.7\,{\pm0.4}$\\
Mudline moment & $22.7\,{\pm0.6}$ & $26.9\,{\pm0.9}$ & $14.1\,{\pm0.6}$ & $44.5\,{\pm32.2}$ & $\mathbf{8.4\,{\pm0.3}}$\\
\cmidrule(l){1-6}
Mean (8 channels) & $16.5\,{\pm0.1}$ & $14.2\,{\pm0.4}$ & $8.7\,{\pm0.2}$ & $17.4\,{\pm9.6}$ & $\mathbf{6.0\,{\pm0.1}}$\\
\bottomrule
\end{tabular*}

\end{table}

\begin{figure}[htbp]
\centering
\begin{subfigure}[t]{\linewidth}\centering
  \panel{a}{\includegraphics{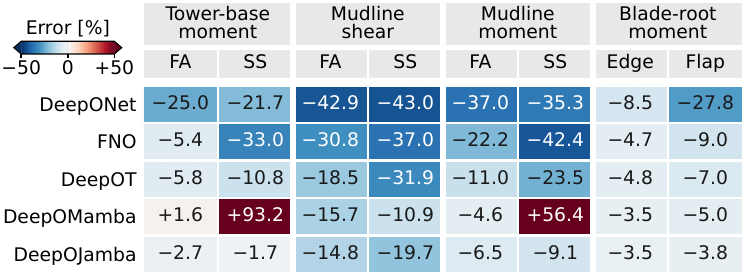}}\phantomsubcaption\label{fig:owt:bias:del}
\end{subfigure}\\[7pt]
\begin{subfigure}[t]{\linewidth}\centering
  \panel{b}{\includegraphics{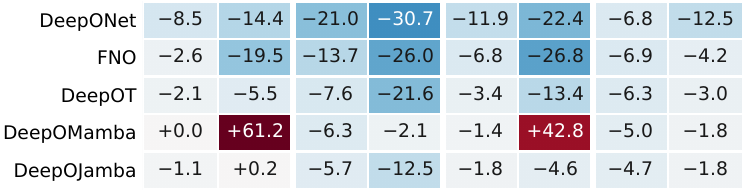}}\phantomsubcaption\label{fig:owt:bias:uls}
\end{subfigure}
\caption{\textbf{Bias of the closed-loop design demands.} Mean signed error $100\,(r-1)$, in percent, of the fatigue \DEL{} (\textbf{a}) and ULS (\textbf{b}) per load channel over the $450$ test simulations. Blue: under-prediction; red: over-prediction; colors saturate at $\pm50\%$. Mean over three training seeds.}
\label{fig:owt:bias}
\end{figure}

Overall, the SS response remains the main source of engineering-demand error for all operators.
The operators differ in the direction of their bias (Fig.~\ref{fig:owt:bias}), and none is uniformly conservative.
DeepOJamba shows less attenuation than DeepOT, FNO, and DeepONet, and its tower-base SS \DEL{} is nearly unbiased (mean signed error $-1.7\%$).
DeepOT generally under-predicts the demands, whereas DeepOMamba over-predicts both SS moments, with mean signed \DEL{} errors of $+93.2\%$ at the tower base and $+56.4\%$ at the mudline.
Lastly, excluding mudline SS shear reduces DeepOJamba's mean \DEL{} and ULS errors to $7.9\%$ and $5.0\%$, respectively, indicating that for most design loads, the proposed operator can give a good design estimate.

\subsection{Sensitivity analysis}

\subsubsection{Effect of wave-heading conditioning on the predicted response}
\label{sec:res:ss}

This section tests whether DeepOJamba is able to adapt its response prediction to the prescribed wave heading direction.
The heading, expressed as the wind--wave misalignment angle $\theta_{\mathrm{mis}}$, is one of the six input parameters used by the simulation descriptor.
The test focuses on the SS loads, which are the most sensitive to $\theta_{\mathrm{mis}}$: under aligned waves they are weakly forced and respond mainly at resonance, whereas under misaligned waves they are forced directly.

\begin{figure}[htbp]
	\centering
	\includegraphics{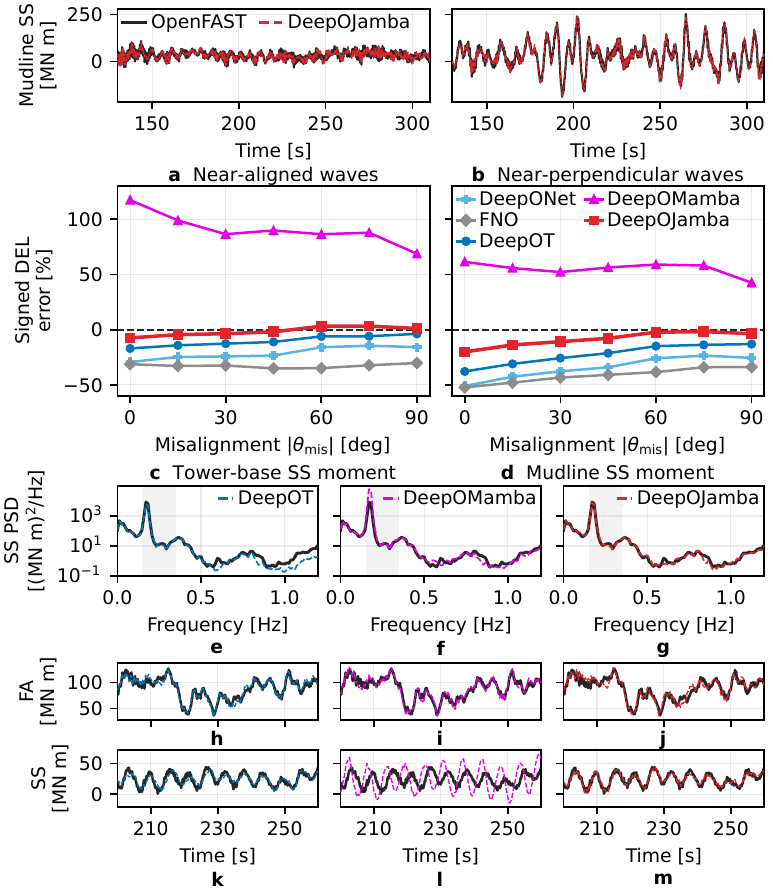}
	{\phantomsubcaption\label{fig:owt:cond:aln}}{\phantomsubcaption\label{fig:owt:cond:perp}}{\phantomsubcaption\label{fig:owt:mis:twr}}{\phantomsubcaption\label{fig:owt:mis:mud}}
	\caption{\textbf{Learned conditioning on wind--wave misalignment and closed-loop tower-base response.} \textbf{a},\textbf{b}, Mudline side--side moment from OpenFAST and DeepOJamba for near-aligned (\textbf{a}) and near-perpendicular (\textbf{b}) waves. \textbf{c},\textbf{d}, Mean signed \DEL{} error of the tower-base (\textbf{c}) and mudline (\textbf{d}) side--side moments per $15^\circ$ misalignment bin, over three training seeds. \textbf{e}--\textbf{m}, Side--side power spectral density, PSD (\textbf{e}--\textbf{g}; resonance band shaded) and tower-base moments (\textbf{h}--\textbf{m}) of one near-aligned test record. FA, fore--aft; SS, side--side.}
	\label{fig:owt:cond}\label{fig:owt:dyn}
\end{figure}

Figures~\ref{fig:owt:cond:aln} and~\ref{fig:owt:cond:perp} compare the responses of two test records with $\theta_{\mathrm{mis}}=5^\circ$ and $79^\circ$.
The remaining parameters ($H_s$, $T_p$, and $\Uhub$) are similar in the two records, so the difference in response is mainly due to the wind-wave misalignment.
In the OpenFAST reference, the s.d. of the mudline SS moment after the start-up transient rises from $24$~MN\,m at $5^\circ$ to $67$~MN\,m at $79^\circ$, and DeepOJamba reproduces both values ($23$ and $67$~MN\,m).
Its signed \DEL{} and ultimate-peak errors are $-9.1\%$ and $+5.9\%$ at $5^\circ$, and $-1.1\%$ and $-4.7\%$ at $79^\circ$.
Thus, the model captures both the change in amplitude and the resulting demands caused by the different misalignment angle.

As shown in Figs.~\ref{fig:owt:mis:twr} and~\ref{fig:owt:mis:mud}, where the \DEL{} error is grouped into bins of the
wind--wave misalignment angle, the accuracy of the SS moments depends strongly on the wave direction.
For DeepOJamba, DeepOT, and DeepONet, a higher wind--wave misalignment angle leads to reduced
error in both trajectory accuracy and engineering design quantities, and their signed \DEL{} errors all move toward zero, each by different amounts.
DeepOJamba stays closest to zero in every bin as its tower-base error changes sign at intermediate misalignment, whereas its mudline error remains negative.
DeepOMamba instead over-predicts both SS moments in every bin, while FNO consistently under-predicts both, with a tower-base error that hardly changes with misalignment.
All operators have their strongest mudline SS shear under-prediction near alignment.

Figures~\ref{fig:owt:dyn}e--g show the power spectral density (PSD) of the tower-base SS moment for DeepOJamba and for the two other sequence operators, DeepOT and DeepOMamba, each compared with the OpenFAST reference.
The grey band marks the $0.15$--$0.35$~Hz range around the first tower mode.
The energy of their SS resonance differs substantially.
Across all $92$ test simulations with $|\theta_{\mathrm{mis}}|<15^\circ$, the median predicted-to-reference energy ratio in the grey band is $1.01$ for DeepOJamba, compared with $0.77$ for DeepOT, which attenuates the resonance, and $10.3$ for DeepOMamba, which amplifies it by about an order of magnitude.
Figures~\ref{fig:owt:dyn}h--m show the corresponding tower-base moment histories in the FA and SS directions.
All three operators track the directly forced FA moment quite well.
For the SS response there are noticeable differences.
DeepOMamba is off-phase and overpredicts the response, while DeepOT follows the phase of the response but underpredicts it. DeepOJamba appears to track the OpenFAST simulation the closest out of all the operators.

\subsubsection{Closed-loop rollout error}
\label{sec:res:chain}

\begin{figure}[htbp]
\centering
\includegraphics{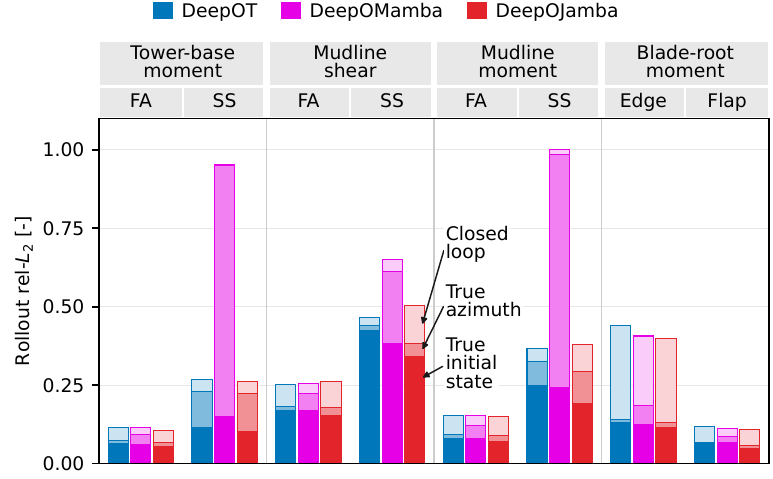}
\caption{\textbf{Decomposition of the closed-loop rollout error.} Post-spin-up \relltwo{} over the $450$ test simulations under end-to-end handoffs, mean over three training seeds. Full bar: closed loop (CL). Dark part: ground-truth-state (GTS) handoffs. Middle part: reduction from reference structural states. Light part: reduction from reference rotor azimuth and speed. \textbf{a}, Per load channel; the FNO blade-root edgewise bar exceeds the axis.}
\label{fig:owt:chain}
\end{figure}

To separate the prediction error within each window from the error propagated through the rollout, the handoff variables predicted by the operator are replaced step by step with the corresponding ground truth values from the OpenFAST simulation.
Three cases are compared in Figure~\ref{fig:owt:chain}:
the fully closed-loop rollout (CL), a rollout in which only the rotor azimuth and speed are taken from the reference ($+$az), and a rollout in which every window starts from the ground-truth state (GTS).
The GTS error is the error made within the windows, and the difference between CL and GTS is the error added by the rollout.

As shown in Figure~\ref{fig:owt:chain} for the three sequential operators, rotor phase accounts for a large amount of the error for DeepOJamba and DeepOMamba.
More specifically, for DeepOJamba, the mean error falls from $0.271$ (CL) to $0.178$ ($+$az) and $0.135$ (GTS).
The rollout therefore adds about $0.14$ to the window error, and about $70\%$ of this excess comes from the rotor phase: correcting the azimuth and rotor speed alone removes about a third of the closed-loop error.
The structural state accounts for the remaining $30\%$ of the excess.
DeepOT and DeepONet follow the same pattern, with the rotor phase again responsible for about $70\%$ of the excess.
DeepOMamba behaves differently. The rotor-phase correction removes only about a tenth of its closed-loop error, whereas the structural state accounts for more than half of it.
FNO lies in between.
The full results are presented in Supplementary Table~4.

Among the $13$ transferred state channels, the two platform velocities have the largest closed-loop errors for most operators.
For the poorly predicted channels, most of the error is phase error (Supplementary Tables~5 and~6 and Supplementary Fig.~3).

A perturbation test of the handed-off SS state (Supplementary Methods - State-error ring-down and handoff interval) shows directly that DeepOMamba does not damp an incoming state error, whereas DeepOJamba and DeepOT do.
A physical perturbation should decay at the operational SS damping rate.
DeepOJamba and DeepOT retain $71\%$ and $83\%$ of this damping, respectively (one training seed).
In DeepOMamba, the perturbation instead grows by $3.5\%$ between the first and sixth tower eigenperiods, corresponding to an equivalent damping of $-0.09\%$.
An incoming error can therefore persist into the next handoff rather than dissipating.
More frequent handoffs strengthen this distinction.
Handing off every quarter of a window raises DeepOJamba's fatigue error by about $40\%$ but more than triples DeepOMamba's, and midpoint handoffs already raise DeepOMamba's error by three quarters (Supplementary Table~2).

\subsubsection{Convolution kernel width}
\label{sec:res:dconv}

The causal convolution inside each Mamba block spans only four time steps by default ($d_{\mathrm{conv}}=4$), that is, in the examined numerical example, $0.04$~s, far less than the tower eigenperiod of $5.65$~s.
To test whether a longer convolution reduces the amplification of state errors, we train both Mamba-based operators, DeepOMamba and the proposed DeepOJamba, with $d_{\mathrm{conv}}=4$, $64$, and $565$ steps.
The widest kernel spans one tower eigenperiod.
The hidden widths are adjusted so that the parameter counts remain comparable.

\begin{figure}[htbp]
\centering
\includegraphics{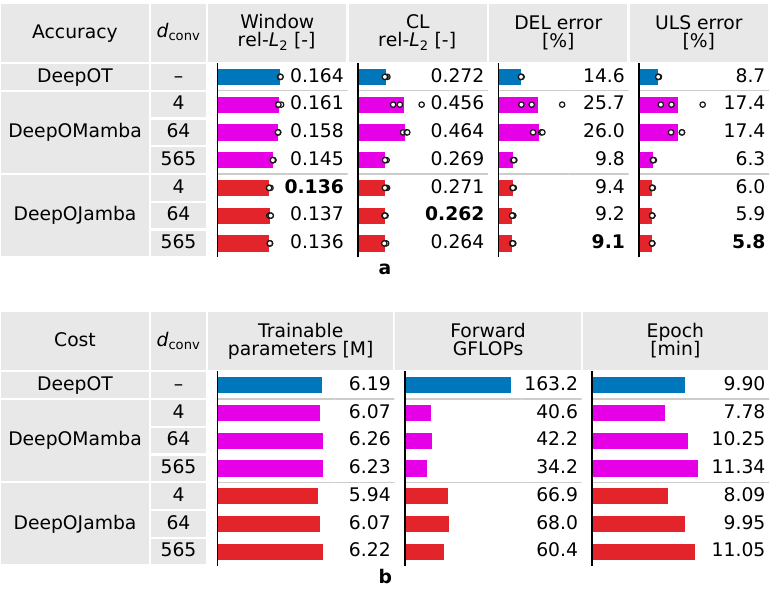}
\caption{\textbf{Sensitivity to the convolution kernel width.} \textbf{a}, Accuracy, as means over the eight loads, with closed-loop (CL) and engineering errors under the end-to-end handoff. Bars and printed values: mean over three training seeds. Dots: the three seeds. Bold: lowest per column. \textbf{b}, Computational cost. FLOPs, floating-point operations of one forward pass over a window.}
\label{fig:owt:dconv}
\end{figure}

The results obtained are summarized in Figure~\ref{fig:owt:dconv} and Supplementary Table~7.
DeepOJamba maintains its engineering accuracy across all three examined widths, with fatigue and ULS errors that improve by only a few percent.
DeepOMamba improves substantially only when the kernel spans a full tower eigenperiod: the $565$-step kernel cuts its fatigue error by about $60\%$.
The improvement occurs primarily during rollout: its window error falls by about $10\%$, but its closed-loop error by about $40\%$.

The wider kernel also restores decay of incoming SS state errors, with equivalent damping reaching $64$--$72\%$ of the physical value across three seeds.
These changes support the interpretation that a period-length kernel limits error amplification in the pure Mamba operator, whereas the attention blocks of DeepOJamba appear to limit it without having to tune the kernel to the structure.

An eigenperiod-length kernel thus lets pure Mamba performance approach the hybrid's, but at a higher computational cost (Section~\ref{sec:res:cost}).
DeepOMamba$_{565}$ still has about $5\%$ and $6\%$ higher fatigue and ULS errors than DeepOJamba, mainly on the tower-base SS moment, which it over-predicts in most simulations while DeepOJamba is nearly unbiased (Supplementary Table~8).
The handoff location can, however, change this comparison (Supplementary Table~2).
DeepOJamba is most accurate with handoffs at the end of the window or at three quarters of it, whereas DeepOMamba$_{565}$ is most accurate at three quarters and degrades less as the handoff moves earlier.

\begin{table}[htbp]
\centering
\caption{\textbf{Closed-loop errors on held-out sea states.} Mean per-simulation error $|r-1|$ in percent over the $450$ held-out wave-band simulations (end-to-end handoff). Mean $\pm$ s.d. over three training seeds. Bold: lowest per row.}
\label{tab:owt:ood}
\setlength{\tabcolsep}{3.5pt}
\setlength{\aboverulesep}{0pt}\setlength{\belowrulesep}{0pt}\renewcommand{\arraystretch}{1.15}
\begin{tabular*}{\linewidth}{@{\extracolsep{\fill}}l cccc|c@{}}
\toprule
Channel & DeepONet & FNO & DeepOT & DeepOMamba & DeepOJamba\\
\midrule
\multicolumn{5}{@{}l|}{\textbf{a}\enspace Fatigue \DEL{} error [\%]} &\\
\multicolumn{5}{@{}l|}{\emph{Fore--aft and blade channels}} &\\
Tower-base moment & $25.1\,{\pm0.2}$ & $7.4\,{\pm0.2}$ & $5.7\,{\pm0.3}$ & $4.9\,{\pm2.0}$ & $\mathbf{3.5\,{\pm0.3}}$\\
Mudline shear & $47.6\,{\pm0.0}$ & $34.7\,{\pm1.7}$ & $20.7\,{\pm0.2}$ & $17.8\,{\pm1.1}$ & $\mathbf{16.7\,{\pm0.5}}$\\
Mudline moment & $38.4\,{\pm0.2}$ & $23.1\,{\pm0.7}$ & $11.1\,{\pm0.4}$ & $6.9\,{\pm1.3}$ & $\mathbf{6.7\,{\pm0.7}}$\\
Blade-root edgewise & $8.4\,{\pm0.2}$ & $4.9\,{\pm0.5}$ & $4.9\,{\pm0.0}$ & $3.6\,{\pm0.6}$ & $\mathbf{3.5\,{\pm0.0}}$\\
Blade-root flapwise & $28.0\,{\pm0.2}$ & $9.7\,{\pm0.9}$ & $7.2\,{\pm0.2}$ & $5.5\,{\pm0.4}$ & $\mathbf{4.4\,{\pm0.1}}$\\
\multicolumn{5}{@{}l|}{\emph{Side--side channels}} &\\
Tower-base moment & $24.4\,{\pm0.5}$ & $33.4\,{\pm1.8}$ & $14.3\,{\pm0.6}$ & $87.1\,{\pm59.6}$ & $\mathbf{12.9\,{\pm0.5}}$\\
Mudline shear & $42.6\,{\pm0.2}$ & $34.8\,{\pm0.3}$ & $29.7\,{\pm0.1}$ & $\mathbf{16.4\,{\pm0.5}}$ & $17.9\,{\pm0.8}$\\
Mudline moment & $35.9\,{\pm0.2}$ & $41.3\,{\pm1.6}$ & $23.5\,{\pm0.8}$ & $52.4\,{\pm39.1}$ & $\mathbf{12.9\,{\pm0.7}}$\\
\cmidrule(l){1-6}
\multicolumn{5}{@{}l|}{Mean (8 channels)} &\\
\quad All ($450$) & $31.3\,{\pm0.0}$ & $23.6\,{\pm0.7}$ & $14.6\,{\pm0.3}$ & $24.3\,{\pm12.9}$ & $\mathbf{9.8\,{\pm0.3}}$\\
\quad Steep strip ($225$) & $28.7\,{\pm0.2}$ & $21.3\,{\pm0.7}$ & $12.0\,{\pm0.2}$ & $22.6\,{\pm11.4}$ & $\mathbf{8.2\,{\pm0.1}}$\\
\quad Long swell ($225$) & $34.0\,{\pm0.1}$ & $26.0\,{\pm0.6}$ & $17.3\,{\pm0.4}$ & $26.1\,{\pm14.5}$ & $\mathbf{11.4\,{\pm0.5}}$\\
\midrule[\heavyrulewidth]
\multicolumn{5}{@{}l|}{\textbf{b}\enspace Ultimate $|$peak$|$ error [\%]} &\\
\multicolumn{5}{@{}l|}{\emph{Fore--aft and blade channels}} &\\
Tower-base moment & $9.3\,{\pm0.3}$ & $6.1\,{\pm0.3}$ & $3.6\,{\pm0.2}$ & $3.3\,{\pm0.1}$ & $\mathbf{3.1\,{\pm0.1}}$\\
Mudline shear & $23.2\,{\pm0.1}$ & $15.3\,{\pm1.0}$ & $8.5\,{\pm0.1}$ & $7.0\,{\pm0.5}$ & $\mathbf{6.5\,{\pm0.4}}$\\
Mudline moment & $12.9\,{\pm0.2}$ & $8.3\,{\pm0.3}$ & $4.2\,{\pm0.1}$ & $3.6\,{\pm0.2}$ & $\mathbf{3.5\,{\pm0.1}}$\\
Blade-root edgewise & $8.0\,{\pm0.2}$ & $7.4\,{\pm0.5}$ & $6.6\,{\pm0.0}$ & $5.3\,{\pm0.8}$ & $\mathbf{4.9\,{\pm0.1}}$\\
Blade-root flapwise & $12.9\,{\pm0.2}$ & $5.5\,{\pm0.1}$ & $3.5\,{\pm0.1}$ & $2.8\,{\pm0.1}$ & $\mathbf{2.6\,{\pm0.0}}$\\
\multicolumn{5}{@{}l|}{\emph{Side--side channels}} &\\
Tower-base moment & $15.8\,{\pm0.4}$ & $20.2\,{\pm1.0}$ & $8.7\,{\pm0.4}$ & $56.7\,{\pm37.4}$ & $\mathbf{8.6\,{\pm0.2}}$\\
Mudline shear & $30.8\,{\pm0.2}$ & $24.9\,{\pm0.4}$ & $20.1\,{\pm0.2}$ & $\mathbf{11.4\,{\pm2.1}}$ & $11.5\,{\pm0.4}$\\
Mudline moment & $23.6\,{\pm0.2}$ & $26.3\,{\pm1.0}$ & $14.0\,{\pm0.4}$ & $39.6\,{\pm28.0}$ & $\mathbf{9.1\,{\pm0.2}}$\\
\cmidrule(l){1-6}
\multicolumn{5}{@{}l|}{Mean (8 channels)} &\\
\quad All ($450$) & $17.1\,{\pm0.1}$ & $14.2\,{\pm0.4}$ & $8.7\,{\pm0.1}$ & $16.2\,{\pm8.6}$ & $\mathbf{6.2\,{\pm0.1}}$\\
\quad Steep strip ($225$) & $16.3\,{\pm0.1}$ & $13.7\,{\pm0.5}$ & $7.5\,{\pm0.1}$ & $15.8\,{\pm7.7}$ & $\mathbf{5.8\,{\pm0.1}}$\\
\quad Long swell ($225$) & $17.8\,{\pm0.1}$ & $14.8\,{\pm0.4}$ & $9.8\,{\pm0.2}$ & $16.6\,{\pm9.6}$ & $\mathbf{6.7\,{\pm0.1}}$\\
\bottomrule
\end{tabular*}

\end{table}

\subsection{Model performance on out-of-distribution sea states}
\label{sec:res:ood}

DeepOJamba is also tested on sea states that it did not see during training.
For this purpose, $450$ additional simulations were generated with wave height and period combinations outside the training range: half with steeper waves and half with long swell.
The wind speeds and misalignment angles stay within the training ranges, so the sea state is the only input that is extrapolated (details in Supplementary Fig.~4 and Supplementary Methods - Environmental loading and dataset).
Table~\ref{tab:owt:ood} compares the fatigue and ultimate-load errors of the five operators for these sea states.
The accuracy of DeepOJamba is almost unchanged, with mean errors that rise by less than half a percentage point compared with the test set, and it remains the most accurate of the five operators.
All five operators are less accurate for long swell than for steep waves.
DeepOJamba therefore extrapolates well to sea states outside its training range, as long as the controller, the turbulence model, and the range of misalignment angles are the same as in training.

\subsection{Computational cost}
\label{sec:res:cost}

Apart from being the most accurate operator, DeepOJamba is also cheaper to train than both DeepOT and DeepOMamba$_{565}$, which are the only other operators of comparable accuracy (Figure~\ref{fig:owt:dconv}b).
With the default kernel width, DeepOJamba needs $41\%$ of the floating-point operations (FLOPs) of DeepOT for one forward pass, and its training epochs are $18\%$ shorter.
Replacing two of DeepOMamba$_4$'s blocks with attention adds only $4\%$ to epoch time but reduces the fatigue error by more than $60\%$.
DeepOMamba$_{565}$ approaches the hybrid's accuracy with about half its forward FLOPs, but requires $40\%$ longer epochs as its long convolution uses a separate fast Fourier transform (FFT) rather than the fused four-step kernel.
Forward FLOPs therefore do not predict the training time here, which depends on how efficiently each operation runs on the graphics processing unit (GPU).

DeepOJamba (7.97GiB) also uses about $13\%$ less training memory than DeepOT (9.28GiB) and about $24\%$ less than DeepOMamba$_{565}$ (10.64GiB).
At inference, it predicts a complete $600$~s record in about half a second, which is as fast as DeepOMamba$_{565}$ and about $10\%$ faster than DeepOT (Supplementary Table~9).

Inference is therefore a negligible part of the surrogate's cost, which is dominated by the generation of its inputs.
Generating the TurbSim wind field and preparing the wave elevation and Airy kinematics takes about $61$~s per record.
The same wind field is also needed by OpenFAST, and its generation is included in the recorded OpenFAST cost of $11.2$ processor core-minutes per simulation.
The surrogate thus replaces only the OpenFAST solution itself, about $10$ core-minutes, with half a second of GPU inference, which gives an overall speed-up of about one order of magnitude per record (hardware in Supplementary Methods - Computational environment and timing).
Generating the inputs in bulk before inference would further increase the speed-up.

\section{Discussion}
\label{sec:discussion}

The OWT benchmark indicates the superb performance of DeepOJamba on lightly damped, multi-modal structural response problems under complex environmental loading.
It also shows where a well-tuned pure Mamba operator remains competitive.
DeepOJamba gives the lowest window, fatigue, and ultimate-load errors of the five operators in the main comparison (Tables~\ref{tab:owt:val} and~\ref{tab:owt:eng}).
It keeps this accuracy without tuning the convolution width and trains faster than both the attention operator and the pure Mamba configuration that comes closest to its engineering accuracy (Fig.~\ref{fig:owt:dconv}).

The two state space operators differ primarily on how they treat an error in the handed-off state during closed-loop rollout.
DeepOJamba reproduces most of the physical decay of an SS state error with its default kernel, whereas
DeepOMamba cannot deliver with the default four-step kernel.
The error persists through the window, and each handoff passes most of it on to the next, so the errors accumulate over the rollout and inflate the SS moments.
The closed-loop error decomposition of Section~\ref{sec:res:chain} confirms that this accumulation, rather than the prediction within each window, is the main source of DeepOMamba's excess error.
Because the four-step kernel spans less than one percent of the $5.65$~s tower eigenperiod, we understand that it leaves the SS mode to the decaying recurrence alone, whereas attention or an eigenperiod-length kernel supplies the window-wide context that regulates it.
Isolating this mechanism inside the network remains an open question.

In the kernel-width study, DeepOJamba keeps its accuracy at every tested width, whereas DeepOMamba approaches it only at the best-tuned width; its errors for different widths were found to exceed two and a half times the hybrid's worst case.
The comparison also depends on the handoff location: DeepOJamba is more accurate than DeepOMamba$_{565}$ with end-to-end handoffs but loses this advantage at handoff points inside the window, most clearly at $25\%$ of the window ($13.3\%$ against $10.2\%$ \DEL{} error; Supplementary Table~2).
Within the tested configurations, DeepOJamba thus removes the need to tune the kernel to the structure. Its drawback is that the error increases when the windows are handed off too frequently.

From an engineering perspective, a surrogate must be judged jointly on the response history and on the design demands derived from it.
A model that reproduces the waveform acceptably can still misestimate fatigue and ultimate loads once its errors accumulate in closed loop or when high-frequency content is present that the operators tend to smooth.
On the turbine, DeepOJamba has the lowest mean window error and retains the best trajectory accuracy with GTS handoffs (Table~\ref{tab:owt:val} and Supplementary Table~4).
Predicted rotor-phase drift nevertheless brings the fully closed-loop \relltwo{} errors of DeepOJamba and DeepOT to similar values ($0.271$ and $0.272$, Fig.~\ref{fig:owt:dconv}a), so better state and phase transfer is needed to carry the hybrid's superior window fidelity into long histories.
Even at this comparable trajectory error, however, DeepOJamba's per-simulation fatigue and ultimate-load errors remain roughly $36\%$ and $31\%$ lower than DeepOT's (Table~\ref{tab:owt:eng}).

The clearest limitation of all the operators is the mudline shear, and the SS component in particular.
Even DeepOJamba has mean per-simulation \DEL{} errors of $14.8\%$ for the FA and $19.7\%$ for the SS mudline shear (Table~\ref{tab:owt:eng}), and it under-predicts both demands systematically.
This deficit is not a rollout effect: it persists when every window starts from the ground-truth initial state (Fig.~\ref{fig:owt:chain}), so it arises within the windows and better state transfer alone will not remove it.
The mudline shear carries more high-frequency content than the moments, and the operators smooth its peaks (Section~\ref{sec:res:window}).

Residual fatigue-demand attenuation also extends beyond shear.
DeepOJamba and DeepOMamba$_{565}$ under-predict all eight and seven of the eight mean \DEL{} ratios, respectively.
Larger models and additional training data, particularly records covering poorly predicted load conditions and independent forcing realizations, could help recover the remaining \DEL{} deficit.
However, the present results do not establish whether capacity, data coverage, architecture, or the window-level training objective is the dominant constraint.
Until these residual biases are resolved, design estimates, especially shear-based ones, require channel-specific calibration or an explicit allowance for under-prediction.
A one-sided conformal prediction bound calibrated on held-out simulations, for example, could offset the systematic model bias\citep{angelopoulos_conformal_2023}.

Three limitations remain to be addressed in future work.
First, errors accumulate during autoregressive rollout because each window starts from interface states that the model itself predicted.
Training over several consecutive windows, in which predicted interface states progressively replace the reference states and are additionally perturbed, would expose the model to its own errors and potentially make it more robust to them \citep{brandstetter_message_2022}.
Enforcing consistency between rotor speed and rotor phase, and between predictions in overlapping windows, could further limit this accumulation.
Second, a single operator currently represents one structure in the power-producing operating regime for three different controller regimes (below-, near-, and above-rated).
A mixture-of-experts architecture could extend this to several regimes by assigning each load case to a specialized sub-network (an expert) based on structural properties, forcing, and controller state.
In this way, one model could represent both linear and nonlinear structural response, as well as power-producing, parked, and faulted turbine operation.
Because only the selected experts are evaluated for each load case, the inference cost would remain low.
Third, DeepOJamba provides a single deterministic prediction without a measure of its uncertainty.
Combining it with a diffusion-based generative model \citep{ho_denoising_2020} would allow several plausible response trajectories to be sampled for each load case, as is done in ensemble weather forecasting \citep{price_probabilistic_2025}.
The spread of these trajectories would quantify the predictive uncertainty, while comparing their mean with held-out simulations would quantify the model bias, so that engineers could judge how far to trust each prediction.
The iterative refinement used by such models has also been shown to extend the accurate rollout of neural partial differential equation solvers \citep{lippe_pde-refiner_2023}.
Finally, sampling could reduce the underprediction of fatigue demands observed in several channels, as a pointwise training loss drives the model to average over response phases that it cannot predict with certainty.
Such model behavior smooths the load cycles and lowers the fatigue demand, whereas sampled trajectories could retain these cycles.

To conclude, DeepOJamba combines state-space recurrence and full attention with physical-parameter conditioning and predicted-state transfer to predict long structural response histories in different controller regimes, accurately enough for both trajectory and engineering-demand assessment.
Such fast surrogates can support applications that require many repeated evaluations, such as design optimization and digital-twin updating.
In online use, however, predictions lag the measurements by one window, because attention needs the forcing over the whole window.
DeepOJamba is a promising backbone for structural foundation models spanning structural families, loading types, and operating regimes, with further gains expected from architectural refinement and larger models and datasets.

\section{Methods}
\label{sec:method}

\subsection{Operator-learning framework}
\label{sec:method:problem}

Dynamic structural-response prediction can be formulated as a map from a forcing history, the state of the structure and its physical properties to the corresponding response history.
Treating the input and output as time-dependent functions allows a model to represent and generalize over many loading histories and parameter combinations \citep{kovachki_neural_2024}.
Let
$\mathbf{f}:[0,T]\to\mathbb{R}^{d_{\mathrm{in}}}$ denote the forcing function and
$\mathbf{u}:[0,T]\to\mathbb{R}^{d_{\mathrm{out}}}$ the structural response, where $d_{\mathrm{in}}$ and $d_{\mathrm{out}}$ are the numbers of input and output channels and $T$ is the sequence duration.
Throughout the paper, bold lowercase symbols denote vectors, bold uppercase symbols denote matrices, and hat denotes model prediction.
We seek a parametric operator $\Gop_\theta$, with trainable parameters $\theta$, whose prediction approximates the response $\mathbf{u}$ and the state trajectory $\mathbf{s}$ of the system.
The operator is defined as,
\begin{equation}
[\widehat{\mathbf{u}},\ \widehat{\mathbf{s}}] \;=\; \Gop_\theta\!\left(\mathbf{f},\,\boldsymbol{\beta},\,\mathbf{s}_0\right)
\label{eq:operator}
\end{equation}
where $\boldsymbol{\beta}\in\mathbb{R}^{d_\beta}$ contains $d_\beta$ structural, loading, or operating descriptors of each realization, and $\mathbf{s}_0=\mathbf{s}(0)\in\mathbb{R}^{d_s}$
is the state at the start of the prediction.

In discrete time, the input is sampled on a uniform grid with $T$ time steps.
Thus, the sampled forcing forms the matrix $\mathbf{F}\in\mathbb{R}^{T\times d_{\mathrm{in}}}$, and the operator predicts $\widehat{\mathbf{u}}\in\mathbb{R}^{T\times d_{\mathrm{out}}}$ and $\widehat{\mathbf{s}}\in\mathbb{R}^{T\times d_s}$.

Within the numerical experiment, the sequence models share the same encoder-decoder design:
\begin{equation}
[\widehat{\mathbf{u}},\ \widehat{\mathbf{s}}] \;=\; \mathrm{Dec}\Big(\,\mathcal{B}_{\boldsymbol{\beta}}\big(\mathrm{Enc}(\mathbf{F},\mathbf{s}_0)\big)\,\Big),
\label{eq:macro}
\end{equation}
where $\mathrm{Enc}$ is the input encoder, $\mathcal{B}_{\boldsymbol{\beta}}$ is a stack of $L$ sequence-mixing backbone blocks conditioned on $\boldsymbol{\beta}$, and $\mathrm{Dec}$ is the output decoder.
The encoder forms one feature vector per time step from the forcing at that step and $\mathbf{s}_0$.
The number of layers, hidden widths, and training settings are listed in Supplementary Tables~3 and~10.

All operators are trained using the batch \relltwo{} loss:
\begin{equation}
\mathcal{L}_{\mathrm{rel}L_2}
=\frac{\big\lVert \widehat{\mathbf{u}} - \mathbf{u}\big\rVert_2}{\big\lVert \mathbf{u}\big\rVert_2},
\label{eq:rell2}
\end{equation}
where $\widehat{\mathbf{u}}$ and $\mathbf{u}$ stack each batch's predicted and target histories, and the norm is taken jointly over the batch, time axis, and output channels.
For a batch of $B$ realizations, let $e_i=\lVert\widehat{\mathbf{u}}_i-\mathbf{u}_i\rVert_2/\lVert \mathbf{u}_i\rVert_2$ be the relative error of realization $i$, and $S_i=\lVert \mathbf{u}_i\rVert_2^{2}$ the squared norm of
its target.
Then
\begin{equation}
\mathcal{L}_{\mathrm{rel}L_2}^{2}
=\frac{\sum_{i=1}^{B}\lVert\widehat{\mathbf{u}}_i-\mathbf{u}_i\rVert_2^{2}}{\sum_{j=1}^{B} S_j}
=\sum_{i=1}^{B} w_i\,e_i^{2},
\qquad
w_i=\frac{S_i}{\sum_{j=1}^{B} S_j},
\label{eq:rell2_weights}
\end{equation}
where $w_i$ is the fraction of target-response energy contributed by realization $i$ and $\sum_i w_i=1$.
The batch ratio therefore weights realizations by their target energy, whereas a mean of per-realization relative errors weights each realization equally.
This reduces the influence of realizations with very small target norms.

Two established operator-learning architectures are examined as baseline comparisons alongside the three sequence models: FNO \citep{li_fourier_2021} and DeepONet \citep{lu_deeponet_2021}.
DeepONet combines a branch network for the input function with a trunk network for the output time $t$, approximating the response as $\sum_{p=1}^{P} b_p\,\tau_p(t)$ with $P$ branch coefficients $b_p$ and trunk basis functions $\tau_p$.
The branch network converts the complete input history, together with $\boldsymbol{\beta}$ and the window-start state, into a set of coefficients.
The trunk network evaluates the basis functions at each requested time and the branch coefficients weight them, so the response is obtained without advancing a structural state from one time step to the next.
FNO parameterizes a kernel-integral operator in the frequency domain.
Each layer multiplies the lowest $k_{\max}$ Fourier modes of the hidden state by learned complex weights and adds a learned pointwise linear term, with $\boldsymbol{\beta}$ and the window-start state concatenated to the input at every step.
Fourier layers provide global, resolution-invariant mixing, but their expressivity is limited by the mode count, as including only global low modes favors smooth, low-frequency responses.
Their turbine configurations are listed in Supplementary Table~10.

The pure sequence operators and the proposed DeepOJamba implement Equation~\eqref{eq:macro} with a per-step encoder, a temporal backbone, and a channel decoder.
They map complete input histories directly to complete response histories over a fixed set of output channels and therefore require no trunk network, which Hu et al.\citep{hu_deepomamba_2025} term sequence-to-sequence operator learning.
Following their \mbox{DeepO-X} convention, DeepOT uses full attention,
DeepOMamba uses Mamba, and the proposed DeepOJamba interleaves a minority
of full-attention layers with a Mamba majority.

\subsubsection{Full-attention operator (DeepOT)}
\label{sec:method:attn}

DeepOT provides the full-attention reference (Supplementary Fig.~5).
For block features $\mathbf{Z}\in\mathbb{R}^{T\times d}$, whose rows are the feature vectors of the $T$ time steps, attention uses content-dependent weights \citep{vaswani_attention_2023}.
With the queries $\mathbf{Q}=\mathbf{Z}\mathbf{W}_Q$, keys $\mathbf{K}=\mathbf{Z}\mathbf{W}_K$, and values $\mathbf{V}=\mathbf{Z}\mathbf{W}_V$ given by learned projections,
\begin{equation}
\mathrm{Attn}(\mathbf{Z})=\mathrm{softmax}\!\Big(\frac{\mathbf{Q}\mathbf{K}^\top}{\sqrt{d_h}}\Big)\mathbf{V},
\label{eq:attn}
\end{equation}
computed in parallel over $H$ heads.
The width $d$ is the feature dimension, which is split across $H$ heads of size
$d_h=d/H$.
Rotary position embeddings supply relative position information
\citep{su_roformer_2023} while no additive positional encoding is used.
Each block also contains pre-normalization with root-mean-square layer normalization (RMSNorm) \citep{zhang_root_2019}, per-head query–key normalization, and a Swish-gated linear unit (SwiGLU) feed-forward sublayer
\citep{shazeer_glu_2020}.

DeepOT stacks $L$ blocks with bidirectional attention over the complete window, giving a leading mixing cost of $\mathcal{O}(LT^2d)$.
All attention blocks in DeepOT and DeepOJamba use exact FlashAttention \citep{dao_flashattention_2022}, which computes attention in tiles. This avoids storing the full $T\times T$ attention matrix and reduces memory movement without changing the attention operation or its quadratic arithmetic.

\subsubsection{Selective state-space operator (DeepOMamba)}
\label{sec:method:mamba}

Whereas full attention directly mixes the complete sequence, DeepOMamba
propagates information through an explicit recurrent state (Supplementary Fig.~6).
This recurrence follows from the continuous linear
state-space model
\begin{equation}
\dot{\mathbf{h}}(t)=\mathbf{A}\,\mathbf{h}(t)+\mathbf{B}\,z(t), \qquad o(t)=\mathbf{C}\,\mathbf{h}(t).
\label{eq:ssm-cont}
\end{equation}
Equation~\eqref{eq:ssm-cont} describes one latent channel.
The scalar input $z(t)$, one channel of the encoded features, updates the internal state $\mathbf{h}(t)\in\mathbb{R}^{d_{\mathrm{state}}}$, and $o(t)$ is the output.
The learned matrices $\mathbf{A}$, $\mathbf{B}$, and $\mathbf{C}$ define state evolution, input injection, and output projection, respectively.
Although the form resembles a structural state equation, $\mathbf{h}$ contains learned features rather than physical displacements, velocities, or constitutive variables.

Holding the input constant over one discretization interval $\Delta$ gives
\begin{equation}
\mathbf{h}_t=\bar{\mathbf{A}}\,\mathbf{h}_{t-1}+\bar{\mathbf{B}}\,z_t, \qquad o_t=\mathbf{C}\,\mathbf{h}_t,
\qquad
\bar{\mathbf{A}}=\exp(\Delta \mathbf{A}), \quad
\bar{\mathbf{B}}=\int_{0}^{\Delta}\exp(\nu\mathbf{A})\mathbf{B}\,\mathrm{d}\nu,
\label{eq:ssm-disc}
\end{equation}
where $t$ indexes time steps, and $\bar{\mathbf{A}},\bar{\mathbf{B}}$ are the discrete state matrices.
The positive model parameter $\Delta$ controls the internal update and is different from the sampling interval $\Delta t$ of the structural record.
Mamba constrains $\mathbf{A}$ to a negative real diagonal, giving decaying homogeneous state components.

Mamba \citep{gu_mamba_2024} makes the recurrence selective by deriving $\mathbf{B}$, $\mathbf{C}$, and $\Delta$ from the encoded input, so that $\bar{\mathbf{A}}_t$, $\bar{\mathbf{B}}_t$, and $\mathbf{C}_t$ in Equation~\eqref{eq:ssm-disc} vary with step $t$.
The step sizes of all channels are
$\boldsymbol{\Delta}_t=\mathrm{softplus}(\mathbf{W}_{\Delta}\tilde{\mathbf{z}}_t+\mathbf{b}_{\Delta})$,
where $\tilde{\mathbf{z}}_t$ is a low-rank projection of the features at step $t$, and $\mathbf{W}_{\Delta}$ and $\mathbf{b}_{\Delta}$ are a learned weight matrix and bias.
All parameters are trained; only the initial range of $\mathrm{softplus}(\mathbf{b}_{\Delta})$; $10^{-3}$–$10^{-1}$ is prescribed (Supplementary Table~10).
Each Mamba layer first applies a causal depthwise convolution of
width $d_{\mathrm{conv}}$ to combine the features of nearby
past steps, after which the selective scan carries information over longer
parts of the loading history.
DeepOMamba stacks $L$ pre-RMSNorm-Mamba-feed-forward blocks, with the feed-forward sublayers mixing information across channels.
The state size $d_{\mathrm{state}}$ sets the number of memory
components available to each channel, while the expansion layer temporarily
widens the representation before returning it to the model width.

The cost of a Mamba block is set by the selective scan and by how the convolution is evaluated.
The selective scan costs $\mathcal{O}(T d\, d_{\mathrm{state}})$.
Direct depthwise convolution adds $\mathcal{O}(Td\,d_{\mathrm{conv}})$.
Zero-padded FFT evaluation computes the same linear convolution in $\mathcal{O}(Td\log T)$ for kernels no longer than the sequence.
On the study GPU, FFT evaluation overtakes direct convolution between widths $64$ and $128$ during training and is about four times faster at width $565$.
The $4$-step kernels use the fused causal-convolution kernel of the Mamba implementation, the $64$-step kernels use direct convolution, and the $565$-step kernels use the FFT, which connects the wider kernels to long-convolution sequence models \citep{poli_hyena_2023,fu_flashfftconv_2024}.

\subsection{Hybrid state-space-attention operator (DeepOJamba)}
\label{sec:method:hybrid}

DeepOJamba interleaves $L_{\mathrm{M}}$ Mamba blocks with $L_{\mathrm{A}}$ full-attention blocks in a single serial backbone (Supplementary Fig.~7).
The construction adapts the Jamba combination principle \citep{lieber_jamba_2024} to sequence-to-sequence operator learning.
For example, writing M for each Mamba block and A for each full-attention block, \texttt{MMAMMAMM} places two attention blocks among six Mamba blocks.
Each symbol includes a complete block with its feed-forward sublayer and residual connections.

Because the complete forcing history is available, attention is bidirectional within each window.
Unlike Jamba, mixture-of-experts layers are omitted to isolate temporal mixing, and routing across different structural response regimes is a possible extension.
At each attention block, information is exchanged across the window before subsequent Mamba blocks propagate the resulting features.
Setting $L_{\mathrm{A}}=0$ reverts to the pure state-space operator.

Attention is added to give every time step direct access to the whole window.
In a Mamba block, the causal convolution weights the features of earlier steps by time lag alone, and only up to a width $d_{\mathrm{conv}}$ fixed in advance.
The selective scan reaches further back: the input $z_s$ at step $s\le t$ enters the output $o_t$ with weight $\mathbf{C}_t\bar{\mathbf{A}}_t\cdots\bar{\mathbf{A}}_{s+1}\bar{\mathbf{B}}_s$ (Equation~\eqref{eq:ssm-disc}), which depends on the content of every step from $s$ to $t$ but passes through $d_{\mathrm{state}}$ components per channel and decays at every step.
In an attention block, the weight between any two steps in the window, earlier or later, depends on their content and relative position, with no width limit, decay, or fixed-size state.
We therefore expect a few attention blocks to help capture dependencies across the window in lightly damped multi-mode responses, whose resonances persist over many cycles, and to reduce the need to tune $d_{\mathrm{conv}}$ to the modal eigenperiods of each structure.

Combining the block costs of Sections~\ref{sec:method:attn} and~\ref{sec:method:mamba}, DeepOJamba costs $\mathcal{O}\big(L_{\mathrm{M}}Td\,(d_{\mathrm{state}}+d_{\mathrm{conv}})+L_{\mathrm{A}}T^{2}d\big)$: the cost remains quadratic in $T$, but only the $L_{\mathrm{A}}$ attention blocks contribute the quadratic term, so the hybrid retains direct global mixing at selected depths at a fraction of DeepOT's attention cost.
We report both forward FLOPs and measured epoch time, because the latter also depends on memory movement, kernel fusion, batching, and hardware.
FLOPs count each convolution as it is evaluated in training, by FFT for the $565$-step kernels; Supplementary Methods - Computational environment and timing describes the epoch-time measurement.

The parameter and initial-state conditioning described below are shared by DeepOJamba, DeepOT, and DeepOMamba.

\subsubsection{Parameter conditioning}
\label{sec:method:cond}

A reusable operator must represent systems with varying structural properties, loading parameters, and operating conditions, collected in $\boldsymbol{\beta}$, and must start from varying initial states $\mathbf{s}_0$.
At each block, we condition using AdaLN-Zero, whose gates are initialized at zero.
A small multilayer perceptron (MLP) maps $\boldsymbol{\beta}$ to its modulation parameters.
This follows its successful use in diffusion transformers \citep{peebles_scalable_2023}.

The parameter network produces a channel-wise scale increment $\tilde{\boldsymbol{\gamma}}$, shift $\boldsymbol{\delta}$, and gate $\boldsymbol{\alpha}$.
Writing $\boldsymbol{\gamma}=\mathbf{1}+\tilde{\boldsymbol{\gamma}}$, the modulation of a feature vector $\mathbf{z}$ is
$\boldsymbol{\gamma}\odot\mathrm{Norm}(\mathbf{z})+\boldsymbol{\delta}$, applied identically at every time step.
Zero-initializing the network's final layer gives $\boldsymbol{\gamma}=\mathbf{1}$ and $\boldsymbol{\delta}=\boldsymbol{\alpha}=\mathbf{0}$.
The gated update is therefore deactivated initially and is learned together with its dependence on $\boldsymbol{\beta}$.
Within each block, the modulation takes one of two placements (Supplementary Fig.~8).
First, the block-level placement conditions the complete mixer--feed-forward block:
\begin{equation}
\mathbf{z}_{\ell+1}
=
(\mathbf{1}-\boldsymbol{\alpha}_\ell)\odot\mathbf{z}_\ell
+\boldsymbol{\alpha}_\ell\odot
\mathrm{Block}_\ell\!\left(
\boldsymbol{\gamma}_\ell\odot\mathrm{Norm}(\mathbf{z}_\ell)+\boldsymbol{\delta}_\ell
\right),
\label{eq:adaln}
\end{equation}
where $\mathbf{z}_\ell$ is the feature vector entering block $\ell$.
Here $\mathrm{Block}_\ell$ includes the mixer's and feed-forward sublayer's internal normalizations and residual connections.
Thus, the modulated features encounter another normalization inside the block.
The channel-wise gate $\boldsymbol{\alpha}_\ell$ combines the input and complete block output while $\boldsymbol{\alpha}_\ell=\mathbf{0}$ passes the input unchanged.

\begin{figure}[htbp]
\centering
\begin{subfigure}{\linewidth}\centering
  \panel{a}{\includegraphics{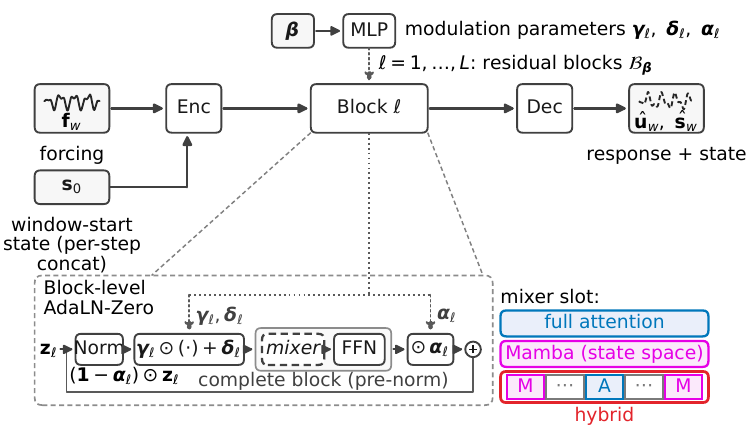}}\phantomsubcaption\label{fig:meth:scaffold}
\end{subfigure}\\[2pt]
\begin{subfigure}{\linewidth}\centering
  \panel{b}{\includegraphics{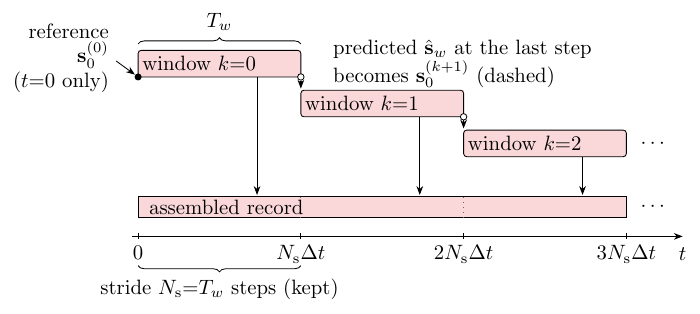}}\phantomsubcaption\label{fig:meth:chain}
\end{subfigure}
\caption{\textbf{Shared sequence-operator architecture and closed-loop chaining.} \textbf{a}, Encoder--decoder design with block-level parameter conditioning and initial-state conditioning; the mixer slot holds full attention (DeepOT), selective state-space blocks (DeepOMamba), or both (DeepOJamba). \textbf{b}, Chaining with end-to-end handoffs: the predicted state $\widehat{\mathbf{s}}_w$ at the last step of window $k$, which is also the first step of window $k+1$, becomes its start state $\mathbf{s}_0^{(k+1)}$.}
\label{fig:meth:arch}
\end{figure}

The sublayer-level placement follows the original
design from Peebles and Xie\citep{peebles_scalable_2023}.
It applies AdaLN-Zero separately to
the block's temporal mixer and feed-forward network (FFN).
The two successive updates are
\begin{equation}
\begin{aligned}
\mathbf{z}^{\mathrm{mix}}_\ell
&= \mathbf{z}_\ell
+ \boldsymbol{\alpha}^{\mathrm{mix}}_\ell \odot
\mathrm{Mixer}_\ell\!\Big(
\boldsymbol{\gamma}^{\mathrm{mix}}_\ell \odot
\widetilde{\mathrm{Norm}}(\mathbf{z}_\ell)
+ \boldsymbol{\delta}^{\mathrm{mix}}_\ell\Big),\\
\mathbf{z}_{\ell+1}
&= \mathbf{z}^{\mathrm{mix}}_\ell
+ \boldsymbol{\alpha}^{\mathrm{ffn}}_\ell \odot
\mathrm{FFN}_\ell\!\Big(
\boldsymbol{\gamma}^{\mathrm{ffn}}_\ell \odot
\widetilde{\mathrm{Norm}}(\mathbf{z}^{\mathrm{mix}}_\ell)
+ \boldsymbol{\delta}^{\mathrm{ffn}}_\ell\Big).
\end{aligned}
\label{eq:adaln_faithful}
\end{equation}
The intermediate features $\mathbf{z}^{\mathrm{mix}}_\ell$ follow the mixer update.
The superscripts $\mathrm{mix}$ and $\mathrm{ffn}$ identify separate modulation and gate vectors, and $\widetilde{\mathrm{Norm}}$ has no learned parameters.
Each branch therefore receives modulation after normalization and contributes its own gated residual.
In a hybrid operator, the placement can be selected separately for Mamba and attention blocks.
DeepOT uses the sublayer-level placement, DeepOMamba the block-level one, and DeepOJamba the block-level placement in its Mamba blocks and the sublayer-level one in its attention blocks.
These placements were chosen based on preliminary experiments (Supplementary Methods - Selection of the conditioning placement and Supplementary Table~11).
Figure~\ref{fig:meth:scaffold} combines the encoder--decoder design of Equation~\eqref{eq:macro} with block-level parameter conditioning; the start state $\mathbf{s}_0$ enters through the encoder (Section~\ref{sec:method:chain}).

\subsubsection{Initial-state conditioning and closed-loop chaining}
\label{sec:method:chain}

Initial-state conditioning enables prediction from nonzero dynamic states and supplies the interface for successive window rollout.
The static descriptor $\boldsymbol{\beta}$ identifies the system or operating condition, while the window-start state $\mathbf{s}_0$ evolves with the response.
We broadcast $\mathbf{s}_0$ along time and concatenate it with the forcing before the encoder; in an ablation, routing it through AdaLN-Zero instead degraded accuracy (Supplementary Methods - Surrogate inputs) for the OWT example.

A long record is predicted by applying Equation~\eqref{eq:operator} to successive windows of $T_w$ steps, as in autoregressive neural solvers that advance several time steps per evaluation and in DeepONet schemes that pass the predicted state at the end of one time interval to the next\citep{brandstetter_message_2022,wang_long-time_2023}.
Window $k$ maps its forcing $\mathbf{f}_w$, $\boldsymbol{\beta}$, and its start state $\mathbf{s}_0^{(k)}$ to $[\widehat{\mathbf{u}}_w,\ \widehat{\mathbf{s}}_w]$, and the predicted state trajectory $\widehat{\mathbf{s}}_w$ supplies the start state of the next window.
Training windows receive reference initial states.
Closed-loop inference uses states predicted by the preceding window.
Only the physical interface state is transferred, with no attention cache or Mamba latent state carried between windows.
The interface must retain the kinematic, constitutive, controller, or phase information needed to continue the response.

At inference, windows are tiled with a stride of $N_{\mathrm{s}}\le T_w-1$ steps, corresponding to an overlap of $1-N_{\mathrm{s}}/T_w$.
Window $k$ begins at step $kN_{\mathrm{s}}$, contributes its first
$N_{\mathrm{s}}$ predictions to the assembled record.
Its prediction at relative step $N_{\mathrm{s}}$, which is the first step of window $k+1$, becomes the start state $\mathbf{s}_0^{(k+1)}$ of window $k+1$.
For exact comparison with the ground truth, if the windows do not divide the record exactly, the last window is shifted back to end at the final step, and only its steps beyond the previous window enter the assembled record. On the turbine, the fifteenth window keeps its last $2{,}670$ steps and starts from the state that the previous window predicts at its start time.
Thus, $N_{\mathrm{s}}=T_w/2$ gives a midpoint handoff and $N_{\mathrm{s}}=T_w-1$ an end-of-window handoff, in which consecutive windows share one step, that is, end-to-end tiling.
Only $\mathbf{s}_0^{(0)}$, the initial state of the complete record, is taken from the reference solution; all later start states are model predictions.
Figure~\ref{fig:meth:chain} illustrates the end-to-end case.
GTS handoffs supply the true state at every handoff, allowing comparison with errors accumulated under predicted-state feedback; on the turbine, $+$az handoffs supply only the true rotor azimuth and speed, so the error reductions attributed to azimuth and to the remaining states depend on this order.

\subsection{Trajectory and engineering error metrics}
\label{sec:method:metrics}

Trajectory accuracy is measured by the \relltwo{} error of the predicted histories in physical units.
The \relltwo{} error is sensitive to both amplitude and phase.
For test record $i$ and response channel $c$,
\begin{equation}
e_{i,c}^{\mathrm{rel}}
=\frac{\left[\sum_{t=1}^{T}
\left(\widehat{u}_{i,c,t}-u_{i,c,t}\right)^2\right]^{1/2}}
{\left[\sum_{t=1}^{T}u_{i,c,t}^{2}\right]^{1/2}},
\label{eq:test_rell2}
\end{equation}
where $u_{i,c,t}$ and $\widehat{u}_{i,c,t}$ are the reference and predicted values of channel $c$
at time step $t$.
The reported channel error and its aggregate over $N_{\mathrm{test}}$ test records and over the $C$ channels are
\begin{equation}
E_c^{\mathrm{rel}}=\frac{1}{N_{\mathrm{test}}}
\sum_{i=1}^{N_{\mathrm{test}}}e_{i,c}^{\mathrm{rel}},
\qquad
E^{\mathrm{rel}}=\frac{1}{C}\sum_{c=1}^{C}E_c^{\mathrm{rel}}.
\label{eq:test_rell2_average}
\end{equation}
Unlike the training loss, these averages weight each record and channel equally.
The window error applies Equation~\eqref{eq:test_rell2} to single windows started from their reference states, $14$ windows of $T_w$ steps per test record with reproducible random starts after the spin-up, so that the averages of Equation~\eqref{eq:test_rell2_average} run over the $6{,}300$ test windows.
The closed-loop error applies it to the post-spin-up part of each assembled record.

We also evaluate peak and cyclic demands, which pointwise accuracy alone does not capture.
For the history $u_c(t)$ of a load channel, rainflow counting
\citep{matsuichi_fatigue_1968,downing_simple_1982} returns the cycle ranges
$\Delta u_j$ and the number of cycles $n_j$ counted at each range.
Under a Wöhler (stress--life) curve with exponent $m$, the Palmgren-Miner damage of these cycles is proportional to $\sum_j n_j\,(\Delta u_j)^{m}$ \citep{miner_cumulative_2021}.
The short-term \DEL{} is the range of a single constant-amplitude load that produces the same damage in a reference number of cycles $N_{\mathrm{ref}}$:
\begin{equation}
\mathrm{DEL} = \left(\frac{\sum_j n_j\,(\Delta u_j)^{m}}{N_{\mathrm{ref}}}\right)^{1/m}.
\label{eq:del}
\end{equation}
Thus, $N_{\mathrm{ref}}$ counts cycles of the one reference load range, the \DEL{} itself, not cycles of the counted history, and $m$ controls the weighting of large cycles.
Because the predicted and reference histories are evaluated with the same $m$ and $N_{\mathrm{ref}}$, the
reference count cancels in their ratio of Equation~\eqref{eq:del_ratio},
\begin{equation}
\frac{\widehat{\mathrm{DEL}}}{\mathrm{DEL}}
=\left(\frac{\sum_j \widehat n_j\,(\Delta\widehat u_j)^{m}}{\sum_j n_j\,(\Delta u_j)^{m}}\right)^{1/m},
\label{eq:del_ratio}
\end{equation}
where each sum runs over the cycles counted in its own history; the predicted and reference histories generally differ in the number and ranges of their cycles.
The value of $m$ used is $m=3$ for the steel tower and monopile channels and $m=10$ for the composite blade roots.

For the ULS, a channel's demand is its peak absolute value over the record, $\max_t |u_c(t)|$.
Both demands are compared per realization as the ratio of predicted to
reference demand, $\widehat q$ and $q$,
\begin{equation}
r=\frac{\widehat q}{q},
\label{eq:demand_ratio}
\end{equation}
so $r=1$ is exact, $r<1$ an under-prediction (non-conservative), and $r>1$
an over-prediction (conservative).
Across $C$ channels and $N_{\mathrm{test}}$ test records, we summarize the error as $E_{\mathrm{cal}}$
\begin{equation}
E_{\mathrm{cal}}=\frac{1}{C}\sum_{c=1}^{C}\frac{1}{N_{\mathrm{test}}}\sum_{i=1}^{N_{\mathrm{test}}}\left|r_{c,i}-1\right|,
\label{eq:calibration_error}
\end{equation}
where $r_{c,i}$ is the ratio of channel $c$ in record $i$.
Taking the absolute deviation of each realization before averaging prevents over- and under-predictions in different realizations from canceling.
The reported error measures the distance from unity regardless of direction, and the sign of the bias is examined separately (Fig.~\ref{fig:owt:bias}).
We report $E_{\mathrm{cal}}$, and the signed deviation $r-1$ where the bias is discussed, in percent.

\subsection{Use of generative artificial intelligence}
\label{sec:method:ai}

Claude (Anthropic) assisted with writing and refactoring the software, including the training, evaluation, and figure-generation scripts and the released code, and with language editing of the manuscript.
The research questions, methods, analyses, and conclusions are the authors' own.
The authors reviewed and tested all artificial-intelligence-assisted code, and every reported number, table, and figure can be computed from the archived evaluation outputs with the released code and was checked against them.
The authors take full responsibility for the content of this article.

\bibliography{sn-bibliography}

\bmhead{Acknowledgements}
The authors gratefully acknowledge the support of the European Union under Project 101169429~--~FOURIER, funded by the Marie Sk\l{}odowska-Curie Doctoral Networks (HORIZON-MSCA-2023-DN-01) within the framework of the European Commission's General Model Grant Agreement.
This work was supported by computational time granted from the National Infrastructures for Research and Technology S.A. (GRNET S.A.) in the National HPC facility~--~ARIS~--~under project ID pr020057\_gpu-TWIN-AI.

\bmhead{Author contributions}
Anastasios Stamou: conceptualization, methodology, software, validation, formal analysis, investigation, visualization, writing (original draft), writing (review and editing).
Rui Teixeira, Maria Nogal, and Michalis Fragiadakis: supervision, conceptualization, formal analysis, writing (original draft), writing (review and editing).

\bmhead{Competing interests}
The authors declare no competing interests.

\bmhead{Supplementary information}
Supplementary Information accompanies this paper: Supplementary Methods, Supplementary Figs.~1--8, and Supplementary Tables~1--11.

\end{document}


\maketitle

\noindent Operator names follow the main text: DeepOT (full attention), DeepOMamba (Mamba), DeepOJamba (hybrid Mamba--attention), FNO (Fourier neural operator), and DeepONet (deep operator network).
Unless a subscript gives another width, DeepOMamba and DeepOJamba use the Mamba default convolution width $d_{\mathrm{conv}}=4$.

\section*{Supplementary Methods}

\subsection*{Turbine model and simulation settings}

The turbine has a nominal rotor diameter of $240$~m and a hub height of $150$~m.
The blade-tip radius is $120.97$~m.
Its steel tower starts at the transition piece, $15$~m above mean sea level, on top of a monopile foundation.
Active degrees of freedom include two fore--aft and two side--side tower modes, the first two flapwise and the first edgewise mode of each blade, generator rotation, nacelle yaw, and all six platform coordinates.
The monopile is represented in SubDyn by a Guyan (zero-internal-mode Craig--Bampton) reduction~\citep{guyan_reduction_1965,craig_coupling_1968} of a $10$~m-diameter steel member in $30$~m of water, rigidly clamped at the mudline without soil--structure interaction.
The irregular wave loading uses Morison strip theory~\citep{morison_force_1950}, without potential-flow, current, second-order, or MacCamy--Fuchs terms; aerodynamics use blade-element momentum theory in AeroDyn.
The integration step is $5$~ms.
The tower and blade modal damping ratios are $1.0\%$ and $0.48\%$, respectively, and the Guyan substructure damping is disabled.
The operational side--side damping is estimated by band-pass random decrement and logarithmic decrement on the tower-base side--side moment of the $450$ test records.

\subsection*{Environmental loading and dataset}

In the International Electrotechnical Commission (IEC)~61400-1 normal turbulence model, class~B sets the reference turbulence intensity $I_{\mathrm{ref}}=0.14$, and the ($90$th-percentile) hub-height turbulence standard deviation grows with wind speed as
\begin{equation}
\sigma_1 \;=\; I_{\mathrm{ref}}\,\bigl(0.75\,\Uhub + 5.6~\mathrm{m/s}\bigr).
\label{eq:owt:ntm}
\end{equation}
Turbulence intensity is therefore not an independent variable; the absolute turbulence rises from $\sigma_1\approx1.1$ to $3.4$~m/s across the sampled range, while the turbulence intensity
$\sigma_1/\Uhub$ falls from ${\sim}37\%$ at the $3$~m/s cut-in to
${\sim}14\%$ at the $25$~m/s cut-out.
Each sampled $\Uhub$ therefore carries the turbulence level that a design-load case at that wind speed would use, rather than an independently drawn one.
The turbulence follows the IEC~61400-1 Kaimal spectrum~\citep{iec61400-1}.
The mean wind follows a power-law shear profile with exponent $0.11$ referenced to the $150$~m hub height.

Inside the moderate-steepness band, $T_p$ is drawn uniformly between bounds that depend on $H_s$,
\begin{equation}
\max\!\bigl(3,\ 3.5\sqrt{H_s}\bigr)\;\le\;T_p\;\le\;\min\!\bigl(18,\ 5\sqrt{H_s}+5\bigr),
\label{eq:owt:tphs}
\end{equation}
with $H_s$ in meters and $T_p$ in seconds.
The lower bound keeps the average wave steepness $S_p=(2\pi/g)\,H_s/T_p^{2}$ below about $1/19$, and the upper bound excludes long swell.
The peak-enhancement factor follows from the DNV-RP-C205 relation
\begin{equation}
\gamma \;=\;
\begin{cases}
5, & T_p/\sqrt{H_s}\le 3.6,\\[2pt]
\exp\!\bigl(5.75-1.15\,T_p/\sqrt{H_s}\bigr), & 3.6<T_p/\sqrt{H_s}<5,\\[2pt]
1, & T_p/\sqrt{H_s}\ge 5,
\end{cases}
\label{eq:owt:gamma}
\end{equation}
so the spectra span peaked JONSWAP (Joint North Sea Wave Project, $\gamma=5$) to the Pierson--Moskowitz~\citep{pierson_proposed_1964} spectral form ($\gamma=1$).

DNV-admissible refers here only to the implemented average-steepness screen:
$S_p\le 1/15$ for $T_p\le8$~s, $S_p\le1/25$ for $T_p\ge15$~s, with the limiting steepness interpolated linearly between these periods.
It does not imply a site-specific environmental contour, a return period, or full design-code compliance.
Under equation~\eqref{eq:owt:gamma}, the held-out steep strip below the training band holds peaked $\gamma=5$ seas and the long-swell band above it $\gamma=1$ seas.

The heading-interpolation cases come from $30$ anchors, each with a fixed wind speed, sea state, TurbSim seed, and wave seed.
Each anchor is simulated at $0^\circ$ and $90^\circ$ for the training pool and at five intermediate headings for testing.
The $150$ heading-interpolation test cases form a controlled test of the misalignment conditioning, because each shares its wind field and wave elevation with training-pool simulations at $0^\circ$ and $90^\circ$ and differs from them only in the heading input.
The $30$ anchors contribute $53$ training and $7$ validation simulations at $0^\circ$ and $90^\circ$, and their test cases use headings of $15^\circ$, $30^\circ$, $45^\circ$, $60^\circ$, and $75^\circ$.
DeepOJamba predicts them with a damage-equivalent load (\DEL{}) error of $8.6\%$, below the $9.8\%$ of the $300$ independent test simulations, and the five operators rank identically on both subsets on the seed means.

The networks operate in a standardized space.
Each of the $73$ OpenFAST output channels is scaled to zero mean and unit variance over simulations and time, the wind field per grid point and velocity component, the derived wave kinematics per component and depth, and the metocean parameters per dimension.
All statistics come from the $1{,}803$ training records alone.
Wave kinematics are evaluated after the first $100$~s, and each evaluated set, whether in-distribution or held out, is standardized using the same statistics.
The loss compares standardized predictions with standardized targets.
At inference, the predictions are mapped back to physical units with the same statistics before the \DEL{} and ultimate limit state (ULS) ratios are computed.

Each simulation starts at the turbine's steady operating point.
Blade pitch and rotor speed follow a ROSCO-tuned schedule interpolated at the mean wind speed, with the same pitch on all three blades: $0^\circ$ below rated, opening to $25^\circ$ at cut-out, while the rotor speed ramps to the rated $7.56$~rpm.
This initialization reduces the start-up transient, and its pitch is the steady-state pitch set-point of the simulation descriptor.

\subsection*{Surrogate inputs}

The two-dimensional wind Convolutional Neural Network (CNN) consists of two stride-$2$ convolutions, $5\times5$ then $3\times3$, with group normalization and Gaussian error linear unit activations.
Among the $21\times21$ TurbSim grid points, we retain only the $293$ points that lie inside the rotor disk (radius $120$~m), and the CNN reduces their three velocity components to $63$ features per time step.
The spatial encoder preserves the grid structure associated with shear, veer, and asymmetric gusts that affect blade and side--side loads.
The $42$ forcing channels are the wave elevation and its time derivative, sixteen Airy velocities and accelerations at four depths down to the $30$~m seabed, sixteen wind-grid summaries, including rotor-equivalent speed, spatial spread, and shear gradients, and $\sin/\cos$ pairs of the once- and three-per-revolution rotor-azimuth harmonics, the instantaneous hub-height wind direction, and its angle to the wave heading.
For DeepOJamba (seed $42$), routing the window-start state through adaptive layer normalization with zero initialization (AdaLN-Zero) instead of per-step concatenation increased the window relative $L_2$ error (\relltwo{}) from $0.135$ to $0.169$ and the \DEL{} error from $9.0\%$ to $15.4\%$, and for DeepOJamba$_{64}$ (seed $42$) from $0.135$ to $0.167$ and from $9.5\%$ to $14.5\%$, supporting the split between the evolving start state and the fixed descriptors.
Both runs that processed the initial state through AdaLN-Zero overfitted: their validation \relltwo{} reached its minimum at epoch $60$ (DeepOJamba) and $65$ (DeepOJamba$_{64}$) of $100$ and then rose while the training loss kept falling; as for every model, the checkpoint with the lowest validation error is used.

\subsection*{Training windows and seeds}

Training samples $14$ windows of $4096$ steps per record per epoch.
With probability $0.25$, a window starts at a random time within the first $100$~s, and otherwise at one of the remaining admissible positions, so the model sees the initial transient and rollouts can start from the solver's prescribed initial state without a warm-up.
The seeds vary the weight initialization, training-window sampling, batch order, and dropout masks.
Each training run seeds PyTorch, CUDA, NumPy, Python, and the data loader; deterministic graphics processing unit (GPU) execution is not enforced, so numerical nondeterminism can add run-to-run variation.
For each run, the checkpoint with the lowest batch-pooled window \relltwo{} error of the standardized outputs on the $257$ validation simulations is retained.

\subsection*{Fatigue-slope sensitivity}

When all eight channels use a common slope $m\in\{3,4,5,10\}$, DeepOJamba retains the lowest mean \DEL{} error ($7.2$--$9.6\%$, compared with $11.2$--$15.0\%$ for DeepOT) and leads on seven channels.
Mudline shear is the most sensitive to the slope.
From $m=3$ to $m=10$, DeepOJamba's mudline-shear \DEL{} error falls from $14.8\%$ to $8.8\%$ in the fore--aft direction and from $19.7\%$ to $13.7\%$ in the side--side direction.
Higher slopes give greater weight to the largest load cycles and generally reduce every operator's \DEL{} error without changing the overall ranking, consistent with much of the \DEL{} error coming from the small, high-frequency cycles that the operators smooth.

\subsection*{State-error ring-down and handoff interval; resonance-band energy}

The ring-down test perturbs the window-start tower-top side--side displacement and velocity by $+0.05$ and by $-0.05$ in normalized units and propagates both perturbed states through the same window with the same prescribed forcing.
The antisymmetric half-difference of the two predicted side--side state trajectories (tower-top displacement and velocity) isolates the response to the perturbation.
Its root-mean-square value over consecutive $565$-step side--side periods (the $5.65$~s tower period), normalized by the first period and taken as the median over $50$ test simulations with four windows each, gives a ring-down whose ratio after six periods, $\rho_6$, defines the equivalent damping $\zeta_{\mathrm{eq}}=-\ln \rho_6/(12\pi)$.
The handoff-interval study uses intervals of $10.24$, $20.48$, $30.72$, and $40.95$~s, that is, handoffs at $25$, $50$, $75$, and $100\%$ of the window (Supplementary Table~\ref{stab:protocol}).
The side--side resonance-band energy ratios use post-spin-up Welch spectra with $81.92$~s segments and $50\%$ overlap.

\subsection*{Computational environment and timing}

Models were trained on an NVIDIA GeForce RTX 5060 Ti GPU or on a multi-instance GPU slice (2g.20gb or 3g.40gb) of an NVIDIA A100-SXM4-80GB GPU of the Greek Research and Technology Network (GRNET) ARIS high-performance computing system, with the same software environment: Python 3.11.14, PyTorch 2.9.1 with CUDA 12.8, and mamba-ssm 2.3.0.
The short-kernel fused-convolution runtime uses causal-conv1d 1.6.2.post1.
Epoch times cover the $2{,}103$ training steps of one epoch (batch $12$, validation excluded) and were measured in one session on the RTX 5060 Ti, with each model at its fastest data-loader setting.
The TurbSim and OpenFAST costs are processor times recorded during dataset generation on an Intel Core Ultra 7 265KF processor.
The rollout times of Supplementary Table~\ref{stab:cost} chain $15$ windows per record end to end on the idle RTX 5060 Ti, with batch size one after one warm-up record and the records read into memory beforehand, include input construction, transfers, predicted-state handoffs, and output assembly, and exclude global input preparation and demand post-processing; the $565$-step kernel is evaluated by fast Fourier transform, as in training.
Training memory is the peak allocated GPU memory of one forward and backward pass with the AdamW update.

\subsection*{Selection of the conditioning placement}
\label{ssec:cond}

The AdaLN-Zero placement of each backbone (Methods) was selected beforehand on a simpler toy problem.
We examined a ten-story nonlinear shear building under synthetic seismic excitation, with story stiffness and yield force varied per record and supplied as the conditioning vector.
Each operator had about $3$\,M parameters (Mamba $d_{\mathrm{conv}}=64$) and was trained on $10^4$ records and tested on $200$.
Modulation inside every block outperformed input concatenation and output-only (late) modulation for every backbone, and among the in-block placements, the block-level one suited DeepOMamba and the sublayer-level one suited DeepOT (Supplementary Table~\ref{stab:cond}).
For DeepOJamba, the mixed placement, which gives each block the placement preferred by its temporal backbone, and the block-level placement differed by less than $4\%$; we kept the mixed placement.
We did not retest the placements on the turbine.

\clearpage
\section*{Supplementary Figures}

\begin{figure}[!htbp]
\centering
\begin{subfigure}{\linewidth}\centering
  \includegraphics{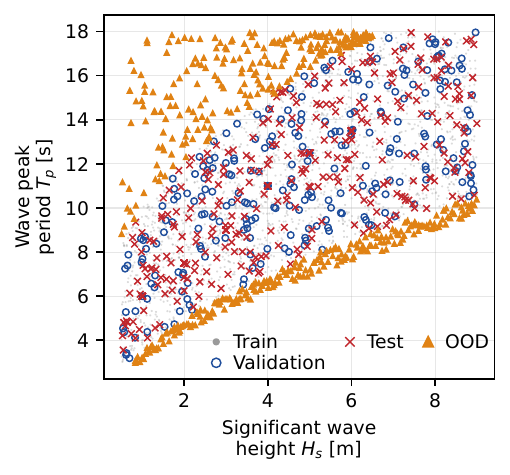}
  \caption{}
\end{subfigure}\\[4pt]
\begin{subfigure}[t]{0.485\linewidth}\centering
  \includegraphics[width=\linewidth]{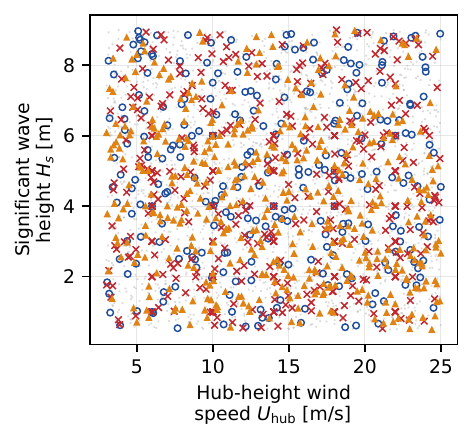}
  \caption{}
\end{subfigure}\hfill
\begin{subfigure}[t]{0.485\linewidth}\centering
  \includegraphics[width=\linewidth]{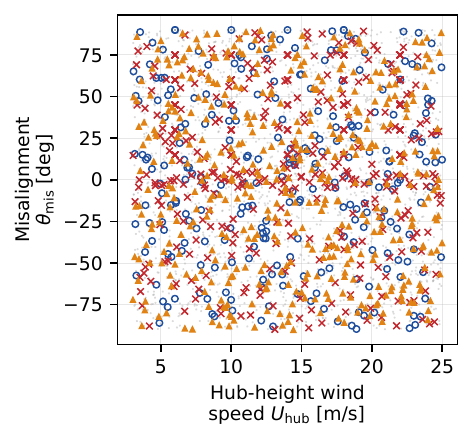}
  \caption{}
\end{subfigure}
\caption{\textbf{Metocean coverage of the four data splits.} \textbf{a}, $T_p$ against $H_s$. \textbf{b}, $H_s$ against $\Uhub$. \textbf{c}, $\theta_{\mathrm{mis}}$ against $\Uhub$. Each simulation appears once: training $1803$, validation $257$, test $450$ and out-of-distribution (OOD) $450$.}
\label{sfig:envcov}
\end{figure}

\begin{figure}[htbp]
\centering
\includegraphics[width=\linewidth]{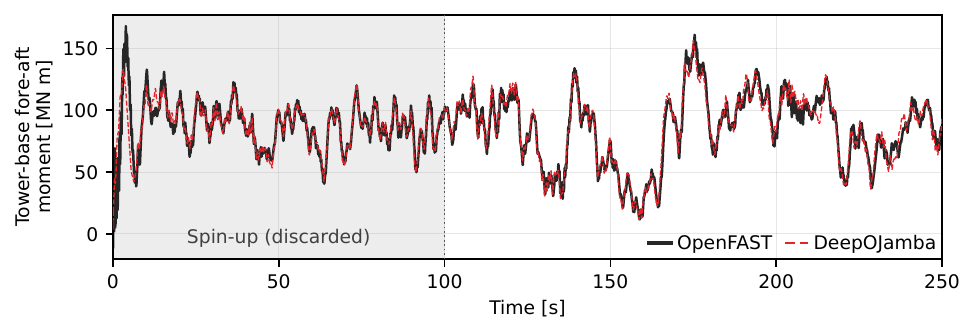}
\caption{\textbf{Closed-loop rollout from the start of a record.} Tower-base fore--aft moment of one test record ($\Uhub\approx17$~m/s, near-aligned waves) predicted from $t=0$ in closed loop (one training seed). Shading: first $100$~s, excluded from all metrics.}
\label{sfig:spinup}
\end{figure}

\begin{figure}[htbp]
\centering
\includegraphics{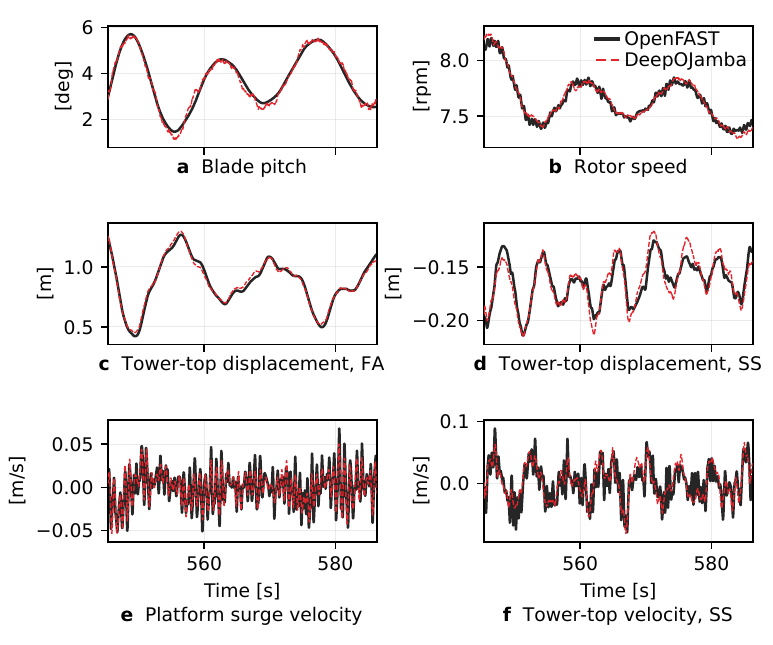}\par\medskip
\begin{subfigure}{0pt}\phantomsubcaption\end{subfigure}\setcounter{subfigure}{6}%
\begin{subfigure}{0.485\linewidth}
  \includegraphics[width=\linewidth]{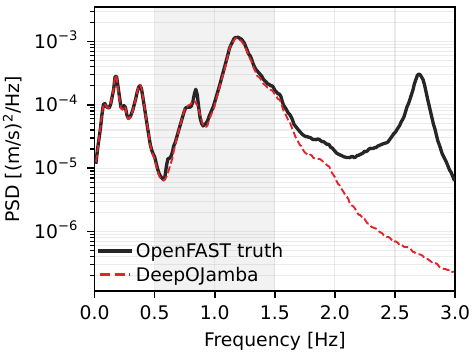}
  \caption{}
\end{subfigure}\hfill
\begin{subfigure}{0.485\linewidth}
  \includegraphics[width=\linewidth]{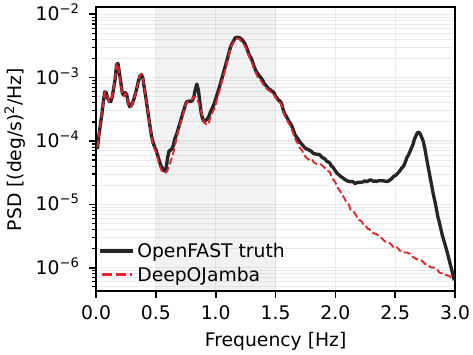}
  \caption{}
\end{subfigure}
\caption{\textbf{Predicted states in one window and platform-velocity spectra in closed loop.} \textbf{a}--\textbf{f}, Six of the thirteen predicted states over the window of main-text Fig.~2a--h, from OpenFAST and DeepOJamba. \textbf{g},\textbf{h}, Mean power spectral density of the platform surge velocity (\textbf{g}) and pitch rate (\textbf{h}) over the $450$ test simulations, from OpenFAST and from DeepOJamba in closed loop. Shading: $0.5$--$1.5$~Hz (most of the variance). FA, fore--aft; SS, side--side; PSD, power spectral density.}
\label{sfig:states}
\end{figure}

\begin{figure}[htbp]
\centering
\begin{subfigure}{0.485\linewidth}
  \includegraphics[width=\linewidth]{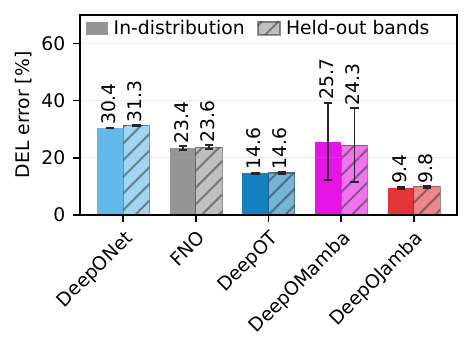}
  \caption{}
\end{subfigure}\hfill
\begin{subfigure}{0.485\linewidth}
  \includegraphics[width=\linewidth]{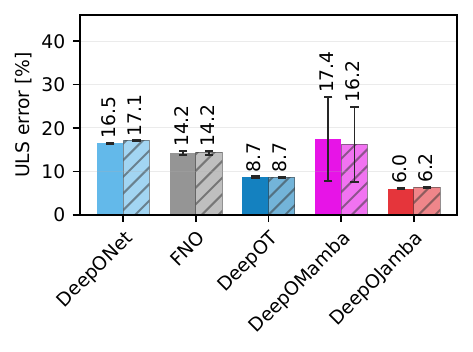}
  \caption{}
\end{subfigure}
\caption{\textbf{Engineering errors in distribution and on the held-out wave bands.} Eight-channel mean per-simulation error $|r-1|$ of the \DEL{} (\textbf{a}) and ULS (\textbf{b}) on the test set and the two held-out wave bands. Bars: mean over three training seeds; whiskers: standard deviation (s.d.).}
\label{sfig:oodcal}
\end{figure}

\begin{figure}[htbp]
\centering
\includegraphics{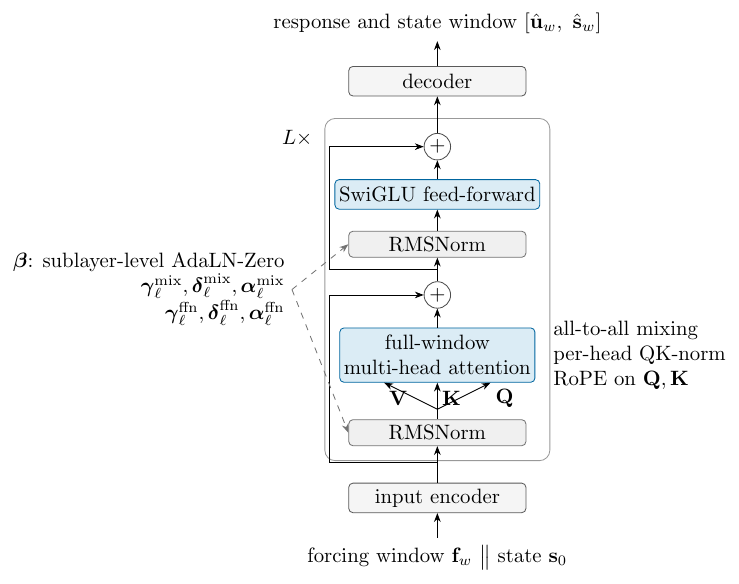}
\caption{\textbf{DeepOT full-attention backbone.} QK-norm, query--key normalization; RMSNorm, root-mean-square layer normalization; RoPE, rotary position embeddings; SwiGLU, Swish-gated linear unit.}
\label{sfig:attn}
\end{figure}

\begin{figure}[htbp]
\centering
\includegraphics{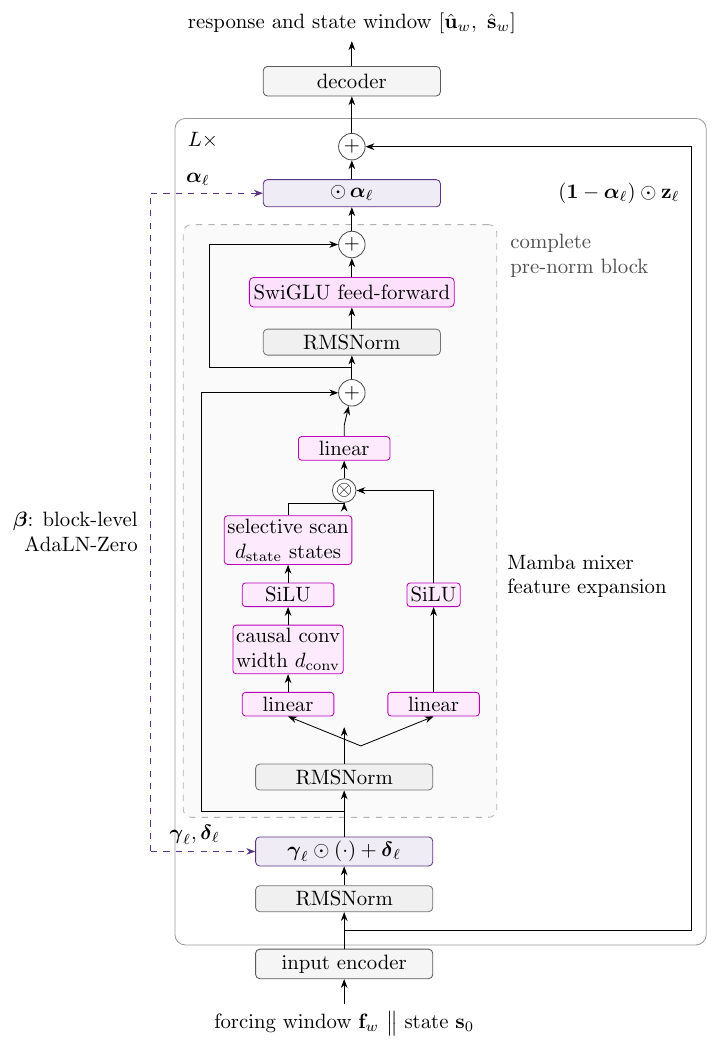}
\caption{\textbf{DeepOMamba selective state-space backbone.} SiLU, sigmoid linear unit.}
\label{sfig:ssm}
\end{figure}

\begin{figure}[htbp]
\centering
\includegraphics{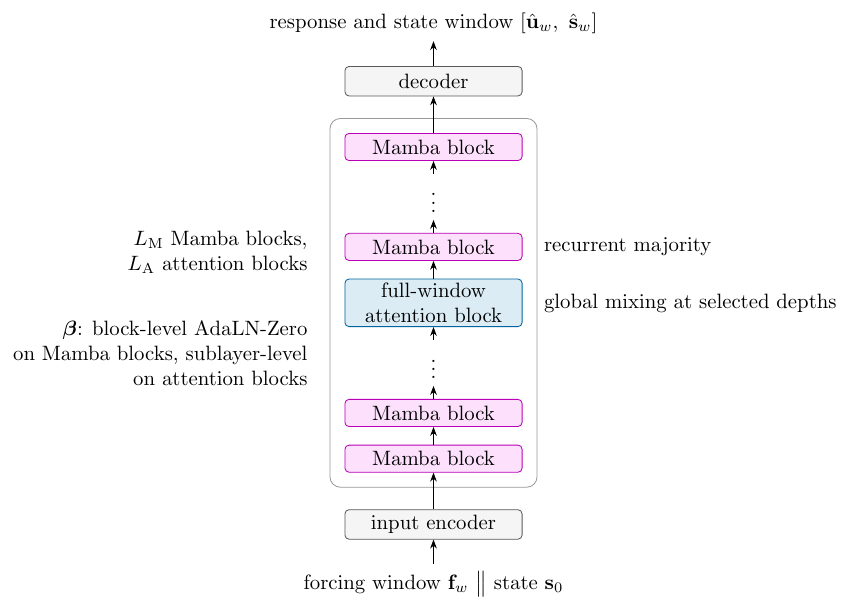}
\caption{\textbf{DeepOJamba interleaved Mamba--attention backbone.}}
\label{sfig:jamba}
\end{figure}

\begin{figure}[htbp]
\centering
\begin{subfigure}{\linewidth}\centering
  \includegraphics{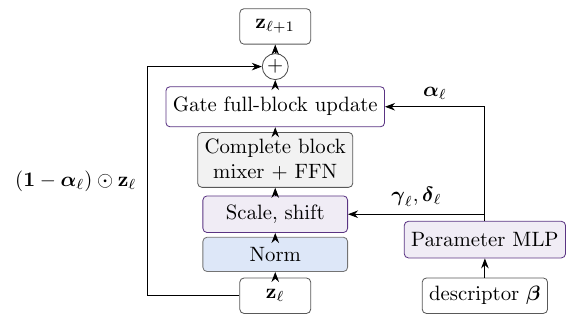}
  \caption{}
\end{subfigure}\\[6pt]
\begin{subfigure}{\linewidth}\centering
  \includegraphics{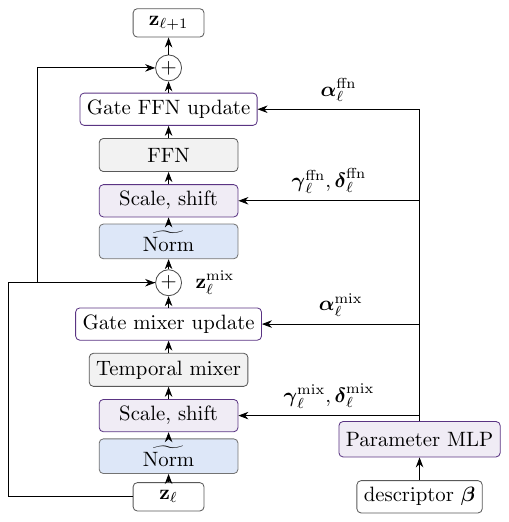}
  \caption{}
\end{subfigure}
\caption{\textbf{AdaLN-Zero placements.} \textbf{a}, Block-level conditioning of the complete mixer and feed-forward network (FFN) block. \textbf{b}, Sublayer-level conditioning of the mixer and FFN residual updates. MLP, multilayer perceptron.}
\label{sfig:cond}
\end{figure}

\clearpage
\section*{Supplementary Tables}
\begin{table}[htbp]
\centering
\caption{\textbf{The $21$ surrogate output channels.} FA, fore--aft; SS, side--side; mudline (SubDyn) channels in N and N\,m, tower and blade (ElastoDyn) channels in kN\,m.}
\label{stab:channels}
\setlength{\tabcolsep}{4.5pt}
\begin{tabular}{l l l}
\toprule
group & Physical meaning & OpenFAST channels\\
\midrule
\multicolumn{3}{l}{\emph{design loads}}\\
tower base & FA and SS bending moments & \texttt{TwrBsMyt}, \texttt{TwrBsMxt}\\
mudline    & FA and SS shears & \texttt{ReactFXss}, \texttt{ReactFYss}\\
mudline    & FA and SS overturning moments & \texttt{ReactMYss}, \texttt{ReactMXss}\\
blade root & flapwise and edgewise moments & \texttt{RootMyb1}, \texttt{RootMxb1}\\
\midrule
\multicolumn{3}{l}{\emph{state channels for chaining}}\\
controller & blade pitches and rotor speed & \texttt{BldPitch1/2/3}, \texttt{RotSpeed}\\
platform   & surge and pitch with their rates & \texttt{PtfmSurge}, \texttt{PtfmTVxi}, \texttt{PtfmPitch}, \texttt{PtfmRVyi}\\
tower FA   & displacement and tower-top velocity & \texttt{TwHt1TPxi}, \texttt{YawBrTVxp}\\
tower SS   & displacement and tower-top velocity & \texttt{TwHt1TPyi}, \texttt{YawBrTVyp}\\
nacelle    & yaw angle & \texttt{NacYaw}\\
\bottomrule
\end{tabular}
\end{table}

\begin{table}[htbp]
\centering
\caption{\textbf{Engineering errors under five handoff protocols.} Eight-channel mean per-simulation error $|r-1|$ in percent over the $450$ test simulations, with the predicted state handed off at $100\%$ (end-to-end), $75\%$, $50\%$ (midpoint) or $25\%$ of the window, or the ground-truth state (GTS) handed off end-to-end. Mean $\pm$ s.d. over three training seeds. Bold: best predicted-state handoff per model.}
\label{stab:protocol}
\begin{tabular}{@{}l ccccc@{}}
\toprule
 & \multicolumn{4}{c}{Predicted state} & GTS\\
\cmidrule(lr){2-5}\cmidrule(l){6-6}
Model & $100\%$ & $75\%$ & $50\%$ & $25\%$ & $100\%$\\
\midrule
\multicolumn{6}{@{}l}{\textbf{a}\enspace Fatigue \DEL{} error [\%]}\\
DeepONet               & $30.4\,{\pm0.1}$ & $29.7\,{\pm0.0}$ & $29.2\,{\pm0.1}$ & $\mathbf{29.1\,{\pm0.3}}$ & $26.8\,{\pm0.1}$\\
FNO                    & $23.4\,{\pm0.7}$ & $19.5\,{\pm0.6}$ & $\mathbf{19.3\,{\pm0.5}}$ & $19.5\,{\pm0.7}$ & $17.5\,{\pm0.4}$\\
DeepOT                 & $14.6\,{\pm0.2}$ & $14.2\,{\pm0.2}$ & $\mathbf{14.0\,{\pm0.1}}$ & $15.1\,{\pm0.3}$ & $12.5\,{\pm0.1}$\\
DeepOMamba             & $\mathbf{25.7\,{\pm13.6}}$ & $32.7\,{\pm8.7}$ & $44.8\,{\pm8.4}$ & $86.2\,{\pm32.5}$ & $8.7\,{\pm0.7}$\\
DeepOMamba$_{565}$     & $9.8\,{\pm0.4}$ & $\mathbf{9.3\,{\pm0.1}}$ & $9.5\,{\pm0.1}$ & $10.2\,{\pm0.1}$ & $7.9\,{\pm0.1}$\\
\cmidrule(l){1-6}
DeepOJamba             & $9.4\,{\pm0.4}$ & $\mathbf{9.3\,{\pm0.3}}$ & $9.8\,{\pm0.3}$ & $13.3\,{\pm0.5}$ & $7.6\,{\pm0.4}$\\
\midrule[\heavyrulewidth]
\multicolumn{6}{@{}l}{\textbf{b}\enspace Ultimate $|$peak$|$ error [\%]}\\
DeepONet               & $16.5\,{\pm0.1}$ & $16.0\,{\pm0.1}$ & $\mathbf{15.7\,{\pm0.0}}$ & $16.0\,{\pm0.2}$ & $14.1\,{\pm0.1}$\\
FNO                    & $14.2\,{\pm0.4}$ & $\mathbf{11.3\,{\pm0.5}}$ & $11.4\,{\pm0.3}$ & $11.6\,{\pm0.4}$ & $9.9\,{\pm0.3}$\\
DeepOT                 & $8.7\,{\pm0.2}$ & $8.5\,{\pm0.1}$ & $\mathbf{8.3\,{\pm0.1}}$ & $9.1\,{\pm0.2}$ & $7.3\,{\pm0.0}$\\
DeepOMamba             & $\mathbf{17.4\,{\pm9.6}}$ & $23.0\,{\pm6.9}$ & $31.9\,{\pm5.5}$ & $55.2\,{\pm16.8}$ & $5.6\,{\pm0.4}$\\
DeepOMamba$_{565}$     & $6.3\,{\pm0.2}$ & $\mathbf{6.0\,{\pm0.0}}$ & $6.1\,{\pm0.2}$ & $6.7\,{\pm0.2}$ & $4.9\,{\pm0.1}$\\
\cmidrule(l){1-6}
DeepOJamba             & $\mathbf{6.0\,{\pm0.1}}$ & $6.1\,{\pm0.1}$ & $6.5\,{\pm0.1}$ & $9.5\,{\pm0.4}$ & $4.7\,{\pm0.1}$\\
\bottomrule
\end{tabular}
\end{table}

\begin{table}[htbp]
\centering
\caption{\textbf{Training hyperparameters.}}
\label{stab:training}
\renewcommand{\arraystretch}{1.15}
\begin{tabular}{p{0.30\linewidth}p{0.55\linewidth}}
\toprule
setting & value\\
\midrule
epochs & $100$\\
batch size & $12$, accumulated over two steps (effective $24$)\\
optimizer & AdamW, $(\beta_1,\beta_2)=(0.9,0.95)$\\
peak learning rate & DeepONet/FNO: $10^{-3}$; DeepOT: $10^{-4}$; DeepOMamba/DeepOJamba: $3\times10^{-4}$\\
weight decay & $10^{-4}$\\
schedule & $5$-epoch linear warmup, then cosine decay to $10^{-3}$ of the peak\\
gradient clipping & $1.0$\\
dropout & $0.15$ (attention blocks, AdaLN-Zero network, baselines); $0.2$ on the mixer and feed-forward outputs of Mamba blocks\\
AdaLN-Zero placement & DeepOT sublayer-level; DeepOMamba block-level; DeepOJamba mixed\\
validation frequency & every $5$ epochs\\
checkpoint criterion & batch-pooled windowed \relltwo{} of the standardized outputs\\
\bottomrule
\end{tabular}
\end{table}

\begin{table}[htbp]
\centering
\caption{\textbf{Decomposition of the closed-loop rollout error.} Post-spin-up \relltwo{} over the $450$ test simulations under closed-loop (CL), true-azimuth ($+$az; reference rotor azimuth and speed) and ground-truth-state (GTS) end-to-end handoffs (mean over three training seeds). Azimuth reduction: CL minus $+$az; state reduction: $+$az minus GTS.}
\label{stab:chain}
\setlength{\tabcolsep}{3.5pt}
\setlength{\aboverulesep}{0pt}\setlength{\belowrulesep}{0pt}\renewcommand{\arraystretch}{1.15}
\begin{tabular*}{\linewidth}{@{\extracolsep{\fill}}l ccccccccc@{}}
\toprule
& \multicolumn{3}{c}{DeepONet} & \multicolumn{3}{c}{FNO} & \multicolumn{3}{c}{DeepOT}\\
\cmidrule(lr){2-4}\cmidrule(lr){5-7}\cmidrule(lr){8-10}
Channel & CL & $+$az & GTS & CL & $+$az & GTS & CL & $+$az & GTS\\
\midrule
\multicolumn{10}{@{}l}{\emph{Fore--aft and blade channels}}\\
Tower-base moment      & 0.154 & 0.142 & 0.128 & 0.198 & 0.171 & 0.085 & 0.115 & 0.074 & 0.065\\
Mudline shear          & 0.302 & 0.301 & 0.297 & 0.251 & 0.242 & 0.229 & 0.252 & 0.183 & 0.169\\
Mudline moment         & 0.172 & 0.164 & 0.154 & 0.202 & 0.178 & 0.120 & 0.152 & 0.094 & 0.081\\
Blade-root edgewise    & 0.836 & 0.213 & 0.207 & 1.424 & 0.392 & 0.155 & 0.439 & 0.139 & 0.130\\
Blade-root flapwise    & 0.215 & 0.167 & 0.164 & 0.377 & 0.172 & 0.094 & 0.119 & 0.067 & 0.062\\
\addlinespace[2pt]
\multicolumn{10}{@{}l}{\emph{Side--side channels}}\\
Tower-base moment      & 0.350 & 0.334 & 0.172 & 0.627 & 0.611 & 0.137 & 0.267 & 0.229 & 0.115\\
Mudline shear          & 0.553 & 0.552 & 0.537 & 0.524 & 0.521 & 0.452 & 0.465 & 0.439 & 0.426\\
Mudline moment         & 0.431 & 0.418 & 0.300 & 0.672 & 0.657 & 0.271 & 0.365 & 0.325 & 0.247\\
\cmidrule(l){1-10}
Mean (8 channels) & 0.377 & 0.286 & 0.245 & 0.534 & 0.368 & 0.193 & 0.272 & 0.194 & 0.162\\
Azimuth reduction & \multicolumn{3}{c}{0.090} & \multicolumn{3}{c}{0.166} & \multicolumn{3}{c}{0.078}\\
State reduction & \multicolumn{3}{c}{0.041} & \multicolumn{3}{c}{0.175} & \multicolumn{3}{c}{0.032}\\
\midrule[\heavyrulewidth]
& \multicolumn{3}{c|}{DeepOMamba} & \multicolumn{3}{c}{DeepOJamba} &  &  & \\
\cmidrule(lr){2-4}\cmidrule(lr){5-7}
Channel & CL & $+$az & \multicolumn{1}{c|}{GTS} & CL & $+$az & GTS &  &  & \\
\midrule
\multicolumn{4}{@{}l|}{\emph{Fore--aft and blade channels}} & & & & & &\\
Tower-base moment      & 0.116 & 0.092 & \multicolumn{1}{c|}{0.061} & 0.106 & 0.066 & 0.054 &  &  & \\
Mudline shear          & 0.256 & 0.222 & \multicolumn{1}{c|}{0.170} & 0.261 & 0.178 & 0.154 &  &  & \\
Mudline moment         & 0.154 & 0.123 & \multicolumn{1}{c|}{0.080} & 0.150 & 0.088 & 0.069 &  &  & \\
Blade-root edgewise    & 0.406 & 0.185 & \multicolumn{1}{c|}{0.124} & 0.400 & 0.130 & 0.115 &  &  & \\
Blade-root flapwise    & 0.112 & 0.086 & \multicolumn{1}{c|}{0.066} & 0.110 & 0.058 & 0.048 &  &  & \\
\multicolumn{4}{@{}l|}{\emph{Side--side channels}} & & & & & &\\
Tower-base moment      & 0.955 & 0.952 & \multicolumn{1}{c|}{0.150} & 0.260 & 0.224 & 0.103 &  &  & \\
Mudline shear          & 0.650 & 0.612 & \multicolumn{1}{c|}{0.383} & 0.503 & 0.384 & 0.342 &  &  & \\
Mudline moment         & 1.002 & 0.985 & \multicolumn{1}{c|}{0.241} & 0.381 & 0.294 & 0.191 &  &  & \\
\cmidrule(l){1-10}
Mean (8 channels) & 0.456 & 0.407 & \multicolumn{1}{c|}{0.159} & 0.271 & 0.178 & 0.135 &  &  & \\
Azimuth reduction & \multicolumn{3}{c|}{0.049} & \multicolumn{3}{c}{0.094} &  &  & \\
State reduction & \multicolumn{3}{c|}{0.248} & \multicolumn{3}{c}{0.043} &  &  & \\
\bottomrule
\end{tabular*}
\end{table}

\begin{table}[htbp]
\centering
\caption{\textbf{Relative $L_2$ error of the state channels.} Closed-loop error (\textbf{a}) over the post-spin-up record and window error (\textbf{b}) over $6{,}300$ windows started from the reference state, for the $13$ state channels of the $450$ test simulations; errors are pooled over simulations or windows because blade pitch is zero below rated wind speed. Mean $\pm$ s.d. over three training seeds. FA, fore--aft; SS, side--side. Bold: lowest per row.}
\label{stab:states}
\footnotesize\setlength{\tabcolsep}{3pt}
\begin{tabular}{@{}l ccccc@{}}
\toprule
State & DeepONet & FNO & DeepOT & DeepOMamba & DeepOJamba\\
\midrule
\multicolumn{6}{@{}l}{\textbf{a}\enspace Closed-loop error (end-to-end handoffs)}\\
Blade 1 pitch                    & $0.0335\,{\pm.0006}$ & $0.0318\,{\pm.0003}$ & $0.0129\,{\pm.0003}$ & $0.0141\,{\pm.0016}$ & $\mathbf{0.0115\,{\pm.0002}}$\\
Blade 2 pitch                    & $0.0335\,{\pm.0006}$ & $0.0318\,{\pm.0003}$ & $0.0129\,{\pm.0003}$ & $0.0142\,{\pm.0017}$ & $\mathbf{0.0115\,{\pm.0002}}$\\
Blade 3 pitch                    & $0.0335\,{\pm.0006}$ & $0.0318\,{\pm.0003}$ & $0.0129\,{\pm.0003}$ & $0.0141\,{\pm.0017}$ & $\mathbf{0.0115\,{\pm.0002}}$\\
Rotor speed                      & $0.0177\,{\pm.0003}$ & $0.0243\,{\pm.0012}$ & $0.0105\,{\pm.0002}$ & $0.0111\,{\pm.0011}$ & $\mathbf{0.0097\,{\pm.0002}}$\\
Platform surge                   & $0.122\,{\pm.001}$ & $0.152\,{\pm.002}$ & $0.104\,{\pm.007}$ & $0.109\,{\pm.009}$ & $\mathbf{0.102\,{\pm.005}}$\\
Platform surge velocity          & $\mathbf{0.985\,{\pm.000}}$ & $1.023\,{\pm.007}$ & $1.122\,{\pm.045}$ & $1.115\,{\pm.004}$ & $1.169\,{\pm.062}$\\
Platform pitch                   & $0.1173\,{\pm.0006}$ & $0.1495\,{\pm.0017}$ & $0.0952\,{\pm.0073}$ & $0.1007\,{\pm.0099}$ & $\mathbf{0.0925\,{\pm.0048}}$\\
Platform pitch rate              & $\mathbf{0.977\,{\pm.001}}$ & $1.035\,{\pm.007}$ & $1.114\,{\pm.053}$ & $1.106\,{\pm.004}$ & $1.159\,{\pm.068}$\\
Tower FA displacement            & $0.1313\,{\pm.0008}$ & $0.1707\,{\pm.0020}$ & $0.0925\,{\pm.0073}$ & $0.1025\,{\pm.0153}$ & $\mathbf{0.0862\,{\pm.0036}}$\\
Tower-top FA velocity            & $0.802\,{\pm.003}$ & $1.051\,{\pm.016}$ & $0.644\,{\pm.077}$ & $0.684\,{\pm.090}$ & $\mathbf{0.609\,{\pm.036}}$\\
Tower SS displacement            & $0.276\,{\pm.002}$ & $0.569\,{\pm.013}$ & $\mathbf{0.203\,{\pm.021}}$ & $0.855\,{\pm.366}$ & $0.205\,{\pm.018}$\\
Tower-top SS velocity            & $0.452\,{\pm.003}$ & $0.890\,{\pm.018}$ & $\mathbf{0.350\,{\pm.034}}$ & $1.316\,{\pm.563}$ & $0.356\,{\pm.029}$\\
Nacelle yaw                      & $0.554\,{\pm.002}$ & $0.665\,{\pm.006}$ & $0.475\,{\pm.054}$ & $0.512\,{\pm.044}$ & $\mathbf{0.474\,{\pm.034}}$\\
\cmidrule(l){1-6}
Mean (13 states)                 & $0.349\,{\pm.001}$ & $0.448\,{\pm.003}$ & $\mathbf{0.327\,{\pm.024}}$ & $0.458\,{\pm.085}$ & $0.331\,{\pm.020}$\\
Mean without platform velocities & $0.234\,{\pm.001}$ & $0.342\,{\pm.002}$ & $0.183\,{\pm.019}$ & $0.339\,{\pm.100}$ & $\mathbf{0.179\,{\pm.012}}$\\
\midrule[\heavyrulewidth]
\multicolumn{6}{@{}l}{\textbf{b}\enspace Window error (reference start states)}\\
Blade 1 pitch                    & $0.0258\,{\pm.0001}$ & $0.0194\,{\pm.0011}$ & $0.0123\,{\pm.0001}$ & $0.0117\,{\pm.0003}$ & $\mathbf{0.0112\,{\pm.0001}}$\\
Blade 2 pitch                    & $0.0258\,{\pm.0001}$ & $0.0194\,{\pm.0011}$ & $0.0123\,{\pm.0001}$ & $0.0117\,{\pm.0002}$ & $\mathbf{0.0112\,{\pm.0001}}$\\
Blade 3 pitch                    & $0.0258\,{\pm.0001}$ & $0.0194\,{\pm.0011}$ & $0.0123\,{\pm.0001}$ & $0.0117\,{\pm.0003}$ & $\mathbf{0.0112\,{\pm.0001}}$\\
Rotor speed                      & $0.0147\,{\pm.0001}$ & $0.0124\,{\pm.0004}$ & $0.0095\,{\pm.0000}$ & $0.0092\,{\pm.0002}$ & $\mathbf{0.0088\,{\pm.0000}}$\\
Platform surge                   & $0.1075\,{\pm.0002}$ & $0.0845\,{\pm.0007}$ & $0.0589\,{\pm.0003}$ & $0.0580\,{\pm.0015}$ & $\mathbf{0.0516\,{\pm.0005}}$\\
Platform surge velocity          & $0.964\,{\pm.000}$ & $0.945\,{\pm.001}$ & $0.605\,{\pm.003}$ & $0.614\,{\pm.008}$ & $\mathbf{0.532\,{\pm.002}}$\\
Platform pitch                   & $0.1014\,{\pm.0003}$ & $0.0770\,{\pm.0008}$ & $0.0557\,{\pm.0003}$ & $0.0544\,{\pm.0015}$ & $\mathbf{0.0491\,{\pm.0006}}$\\
Platform pitch rate              & $0.946\,{\pm.000}$ & $0.911\,{\pm.001}$ & $0.512\,{\pm.004}$ & $0.524\,{\pm.010}$ & $\mathbf{0.425\,{\pm.002}}$\\
Tower FA displacement            & $0.1102\,{\pm.0006}$ & $0.0791\,{\pm.0016}$ & $0.0639\,{\pm.0004}$ & $0.0612\,{\pm.0018}$ & $\mathbf{0.0573\,{\pm.0008}}$\\
Tower-top FA velocity            & $0.677\,{\pm.001}$ & $0.385\,{\pm.006}$ & $0.298\,{\pm.003}$ & $0.291\,{\pm.008}$ & $\mathbf{0.254\,{\pm.004}}$\\
Tower SS displacement            & $0.1333\,{\pm.0003}$ & $0.1028\,{\pm.0020}$ & $0.0840\,{\pm.0016}$ & $0.1175\,{\pm.0127}$ & $\mathbf{0.0750\,{\pm.0019}}$\\
Tower-top SS velocity            & $0.256\,{\pm.000}$ & $0.212\,{\pm.002}$ & $0.178\,{\pm.002}$ & $0.215\,{\pm.017}$ & $\mathbf{0.157\,{\pm.003}}$\\
Nacelle yaw                      & $0.520\,{\pm.000}$ & $0.359\,{\pm.004}$ & $0.250\,{\pm.003}$ & $0.272\,{\pm.002}$ & $\mathbf{0.204\,{\pm.005}}$\\
\cmidrule(l){1-6}
Mean (13 states)                 & $0.301\,{\pm.000}$ & $0.248\,{\pm.002}$ & $0.165\,{\pm.001}$ & $0.173\,{\pm.005}$ & $\mathbf{0.142\,{\pm.001}}$\\
Mean without platform velocities & $0.1815\,{\pm.0003}$ & $0.1245\,{\pm.0019}$ & $0.0941\,{\pm.0008}$ & $0.1012\,{\pm.0039}$ & $\mathbf{0.0809\,{\pm.0014}}$\\
\bottomrule
\end{tabular}
\end{table}

\begin{table}[htbp]
\centering
\caption{\textbf{Root-mean-square error of the state channels.} Closed-loop error (\textbf{a}) and window error (\textbf{b}) of the $13$ state channels in the units listed (mm, mm/s and $10^{-3}$\,deg are scaled OpenFAST units), evaluated as in Supplementary Table~\ref{stab:states}. Mean $\pm$ s.d. over three training seeds. FA, fore--aft; SS, side--side. Bold: lowest per row.}
\label{stab:states_rmse}
\footnotesize\setlength{\tabcolsep}{3pt}
\begin{tabular}{@{}l l ccccc@{}}
\toprule
State & Unit & DeepONet & FNO & DeepOT & DeepOMamba & DeepOJamba\\
\midrule
\multicolumn{7}{@{}l}{\textbf{a}\enspace Closed-loop error (end-to-end handoffs)}\\
Blade 1 pitch                    & deg & $0.420\,{\pm.007}$ & $0.399\,{\pm.003}$ & $0.161\,{\pm.004}$ & $0.177\,{\pm.020}$ & $\mathbf{0.145\,{\pm.002}}$\\
Blade 2 pitch                    & deg & $0.420\,{\pm.007}$ & $0.399\,{\pm.003}$ & $0.161\,{\pm.004}$ & $0.178\,{\pm.021}$ & $\mathbf{0.144\,{\pm.002}}$\\
Blade 3 pitch                    & deg & $0.420\,{\pm.007}$ & $0.399\,{\pm.003}$ & $0.162\,{\pm.004}$ & $0.177\,{\pm.021}$ & $\mathbf{0.145\,{\pm.002}}$\\
Rotor speed                      & rpm & $0.1237\,{\pm.0023}$ & $0.1693\,{\pm.0084}$ & $0.0735\,{\pm.0015}$ & $0.0776\,{\pm.0074}$ & $\mathbf{0.0678\,{\pm.0014}}$\\
Platform surge                   & mm & $5.19\,{\pm.02}$ & $6.48\,{\pm.07}$ & $4.41\,{\pm.32}$ & $4.61\,{\pm.37}$ & $\mathbf{4.35\,{\pm.23}}$\\
Platform surge velocity          & mm/s & $\mathbf{20.4\,{\pm.0}}$ & $21.2\,{\pm.1}$ & $23.3\,{\pm.9}$ & $23.1\,{\pm.1}$ & $24.2\,{\pm1.3}$\\
Platform pitch                   & $10^{-3}$\,deg & $12.27\,{\pm.06}$ & $15.64\,{\pm.18}$ & $9.96\,{\pm.76}$ & $10.53\,{\pm1.04}$ & $\mathbf{9.68\,{\pm.51}}$\\
Platform pitch rate              & $10^{-3}$\,deg/s & $\mathbf{37.6\,{\pm.0}}$ & $39.8\,{\pm.3}$ & $42.9\,{\pm2.0}$ & $42.5\,{\pm.2}$ & $44.6\,{\pm2.6}$\\
Tower FA displacement            & mm & $62.1\,{\pm.4}$ & $80.7\,{\pm.9}$ & $43.7\,{\pm3.4}$ & $48.4\,{\pm7.2}$ & $\mathbf{40.8\,{\pm1.7}}$\\
Tower-top FA velocity            & mm/s & $67.8\,{\pm.3}$ & $88.7\,{\pm1.4}$ & $54.4\,{\pm6.5}$ & $57.8\,{\pm7.6}$ & $\mathbf{51.5\,{\pm3.0}}$\\
Tower SS displacement            & mm & $56.2\,{\pm.4}$ & $115.8\,{\pm2.6}$ & $\mathbf{41.3\,{\pm4.2}}$ & $174.1\,{\pm74.6}$ & $41.8\,{\pm3.7}$\\
Tower-top SS velocity            & mm/s & $69.0\,{\pm.4}$ & $135.8\,{\pm2.7}$ & $\mathbf{53.5\,{\pm5.2}}$ & $200.9\,{\pm86.0}$ & $54.3\,{\pm4.4}$\\
Nacelle yaw                      & $10^{-3}$\,deg & $43.7\,{\pm.2}$ & $52.5\,{\pm.4}$ & $37.5\,{\pm4.3}$ & $40.4\,{\pm3.5}$ & $\mathbf{37.4\,{\pm2.7}}$\\
\midrule[\heavyrulewidth]
\multicolumn{7}{@{}l}{\textbf{b}\enspace Window error (reference start states)}\\
Blade 1 pitch                    & deg & $0.322\,{\pm.002}$ & $0.242\,{\pm.014}$ & $0.154\,{\pm.001}$ & $0.147\,{\pm.003}$ & $\mathbf{0.140\,{\pm.001}}$\\
Blade 2 pitch                    & deg & $0.322\,{\pm.002}$ & $0.242\,{\pm.014}$ & $0.154\,{\pm.001}$ & $0.147\,{\pm.003}$ & $\mathbf{0.140\,{\pm.001}}$\\
Blade 3 pitch                    & deg & $0.322\,{\pm.002}$ & $0.242\,{\pm.014}$ & $0.154\,{\pm.001}$ & $0.146\,{\pm.003}$ & $\mathbf{0.140\,{\pm.001}}$\\
Rotor speed                      & rpm & $0.1028\,{\pm.0004}$ & $0.0867\,{\pm.0025}$ & $0.0662\,{\pm.0003}$ & $0.0640\,{\pm.0012}$ & $\mathbf{0.0617\,{\pm.0001}}$\\
Platform surge                   & mm & $4.57\,{\pm.01}$ & $3.59\,{\pm.03}$ & $2.50\,{\pm.01}$ & $2.47\,{\pm.06}$ & $\mathbf{2.20\,{\pm.02}}$\\
Platform surge velocity          & mm/s & $20.0\,{\pm.0}$ & $19.6\,{\pm.0}$ & $12.6\,{\pm.1}$ & $12.8\,{\pm.2}$ & $\mathbf{11.1\,{\pm.0}}$\\
Platform pitch                   & $10^{-3}$\,deg & $10.62\,{\pm.03}$ & $8.07\,{\pm.09}$ & $5.84\,{\pm.03}$ & $5.70\,{\pm.15}$ & $\mathbf{5.15\,{\pm.06}}$\\
Platform pitch rate              & $10^{-3}$\,deg/s & $36.5\,{\pm.0}$ & $35.1\,{\pm.0}$ & $19.7\,{\pm.2}$ & $20.2\,{\pm.4}$ & $\mathbf{16.4\,{\pm.1}}$\\
Tower FA displacement            & mm & $52.2\,{\pm.3}$ & $37.4\,{\pm.8}$ & $30.3\,{\pm.2}$ & $29.0\,{\pm.9}$ & $\mathbf{27.1\,{\pm.4}}$\\
Tower-top FA velocity            & mm/s & $57.4\,{\pm.1}$ & $32.6\,{\pm.5}$ & $25.2\,{\pm.2}$ & $24.7\,{\pm.7}$ & $\mathbf{21.5\,{\pm.3}}$\\
Tower SS displacement            & mm & $27.4\,{\pm.1}$ & $21.1\,{\pm.4}$ & $17.2\,{\pm.3}$ & $24.1\,{\pm2.6}$ & $\mathbf{15.4\,{\pm.4}}$\\
Tower-top SS velocity            & mm/s & $39.5\,{\pm.0}$ & $32.7\,{\pm.4}$ & $27.5\,{\pm.3}$ & $33.3\,{\pm2.6}$ & $\mathbf{24.3\,{\pm.5}}$\\
Nacelle yaw                      & $10^{-3}$\,deg & $41.3\,{\pm.0}$ & $28.5\,{\pm.3}$ & $19.9\,{\pm.2}$ & $21.6\,{\pm.1}$ & $\mathbf{16.2\,{\pm.4}}$\\
\bottomrule
\end{tabular}
\end{table}

\begin{table}[htbp]
\centering
\caption{\textbf{Sensitivity to the convolution kernel width.} Errors are means over the eight loads, with closed-loop (CL) and engineering errors under the end-to-end handoff; FLOPs, floating-point operations of one forward pass over a window. Accuracy columns: mean $\pm$ s.d. over three training seeds. Bold: lowest per column.}
\label{stab:dconv}
\setlength{\tabcolsep}{3pt}
\begin{tabular*}{\linewidth}{@{\extracolsep{\fill}}l c cccc@{}}
\toprule
\textbf{a}\enspace Accuracy & $d_{\mathrm{conv}}$ & \makecell{Window\\\relltwo{}} & \makecell{CL\\\relltwo{}} & \makecell{\DEL{} error\\{}[\%]} & \makecell{ULS error\\{}[\%]}\\
\midrule
DeepOT & -- & $0.164\,{\pm.001}$ & $0.272\,{\pm.013}$ & $14.6\,{\pm0.2}$ & $8.7\,{\pm0.2}$\\
\addlinespace[2pt]
DeepOMamba & $4$ & $0.161\,{\pm.004}$ & $0.456\,{\pm.145}$ & $25.7\,{\pm13.6}$ & $17.4\,{\pm9.6}$\\
DeepOMamba & $64$ & $0.158\,{\pm.000}$ & $0.464\,{\pm.022}$ & $26.0\,{\pm3.3}$ & $17.4\,{\pm2.8}$\\
DeepOMamba & $565$ & $0.145\,{\pm.001}$ & $0.269\,{\pm.008}$ & $9.8\,{\pm0.4}$ & $6.3\,{\pm0.2}$\\
\cmidrule(l){1-6}
DeepOJamba & $4$ & $\mathbf{0.136\,{\pm.002}}$ & $0.271\,{\pm.011}$ & $9.4\,{\pm0.4}$ & $6.0\,{\pm0.1}$\\
DeepOJamba & $64$ & $0.137\,{\pm.002}$ & $\mathbf{0.262\,{\pm.006}}$ & $9.2\,{\pm0.4}$ & $5.9\,{\pm0.0}$\\
DeepOJamba & $565$ & $0.136\,{\pm.000}$ & $0.264\,{\pm.009}$ & $\mathbf{9.1\,{\pm0.4}}$ & $\mathbf{5.8\,{\pm0.1}}$\\
\midrule[\heavyrulewidth]
\textbf{b}\enspace Computational cost & $d_{\mathrm{conv}}$ & \makecell{Params\\{}[M]} & \makecell{Forward\\GFLOPs} & \makecell{Epoch\\{}[min]} &\\
\midrule
DeepOT & -- & $6.19$ & $163.2$ & $9.90$ &\\
\addlinespace[2pt]
DeepOMamba & $4$ & $6.07$ & $40.6$ & $7.78$ &\\
DeepOMamba & $64$ & $6.26$ & $42.2$ & $10.25$ &\\
DeepOMamba & $565$ & $6.23$ & $34.2$ & $11.34$ &\\
\cmidrule(l){1-6}
DeepOJamba & $4$ & $5.94$ & $66.9$ & $8.09$ &\\
DeepOJamba & $64$ & $6.07$ & $68.0$ & $9.95$ &\\
DeepOJamba & $565$ & $6.22$ & $60.4$ & $11.05$ &\\
\bottomrule
\end{tabular*}
\end{table}

\begin{table}[htbp]
\centering
\caption{\textbf{Per-channel engineering errors of the convolution-width study.} Mean per-simulation error $|r-1|$ in percent over the $450$ test simulations (end-to-end handoff). Mean $\pm$ s.d. over three training seeds. Bold: lowest per row.}
\label{stab:kernelch}
\small\setlength{\tabcolsep}{3.5pt}
\setlength{\aboverulesep}{0pt}\setlength{\belowrulesep}{0pt}\renewcommand{\arraystretch}{1.15}
\begin{tabular}{@{}l ccc|ccc@{}}
\toprule
& \multicolumn{3}{c|}{DeepOMamba, $d_{\mathrm{conv}}$} & \multicolumn{3}{c}{DeepOJamba, $d_{\mathrm{conv}}$}\\
\cmidrule(lr){2-4}\cmidrule(lr){5-7}
Channel & $4$ & $64$ & $565$ & $4$ & $64$ & $565$\\
\midrule
\multicolumn{4}{@{}l|}{\textbf{a}\enspace Fatigue \DEL{} error [\%]} & & &\\
\multicolumn{4}{@{}l|}{\emph{Fore--aft and blade channels}} & & &\\
Tower-base moment            & $4.6{\pm1.9}$ & $4.7{\pm0.9}$ & $\mathbf{2.5{\pm0.2}}$ & $3.2{\pm0.2}$ & $3.0{\pm0.4}$ & $3.3{\pm0.4}$\\
Mudline shear                & $15.7{\pm0.9}$ & $16.1{\pm0.3}$ & $15.6{\pm0.1}$ & $14.8{\pm0.5}$ & $\mathbf{14.5{\pm0.6}}$ & $\mathbf{14.5{\pm0.5}}$\\
Mudline moment               & $6.5{\pm1.0}$ & $6.9{\pm0.6}$ & $6.4{\pm0.1}$ & $6.5{\pm0.7}$ & $\mathbf{6.1{\pm0.8}}$ & $6.2{\pm0.8}$\\
Blade-root edgewise          & $3.6{\pm0.6}$ & $4.1{\pm0.2}$ & $3.8{\pm0.1}$ & $3.5{\pm0.0}$ & $\mathbf{3.3{\pm0.2}}$ & $3.4{\pm0.3}$\\
Blade-root flapwise          & $5.3{\pm0.3}$ & $5.3{\pm0.1}$ & $5.0{\pm0.1}$ & $4.0{\pm0.1}$ & $\mathbf{3.9{\pm0.6}}$ & $4.2{\pm0.2}$\\
\multicolumn{4}{@{}l|}{\emph{Side--side channels}} & & &\\
Tower-base moment            & $93.7{\pm62.5}$ & $95.6{\pm13.5}$ & $14.5{\pm2.9}$ & $11.3{\pm0.6}$ & $11.5{\pm0.9}$ & $\mathbf{9.6{\pm0.2}}$\\
Mudline shear                & $16.9{\pm0.7}$ & $\mathbf{15.7{\pm0.5}}$ & $19.2{\pm0.8}$ & $19.7{\pm0.9}$ & $19.7{\pm0.7}$ & $19.9{\pm0.9}$\\
Mudline moment               & $58.9{\pm42.5}$ & $59.4{\pm10.7}$ & $11.5{\pm0.5}$ & $12.0{\pm1.1}$ & $\mathbf{11.3{\pm0.9}}$ & $11.8{\pm0.5}$\\
\cmidrule(l){1-7}
Mean (8 channels)            & $25.7{\pm13.6}$ & $26.0{\pm3.3}$ & $9.8{\pm0.4}$ & $9.4{\pm0.4}$ & $9.2{\pm0.4}$ & $\mathbf{9.1{\pm0.4}}$\\
\midrule[\heavyrulewidth]
\multicolumn{4}{@{}l|}{\textbf{b}\enspace Ultimate $|$peak$|$ error [\%]} & & &\\
\multicolumn{4}{@{}l|}{\emph{Fore--aft and blade channels}} & & &\\
Tower-base moment            & $2.9{\pm0.2}$ & $2.9{\pm0.1}$ & $\mathbf{2.6{\pm0.1}}$ & $\mathbf{2.6{\pm0.1}}$ & $\mathbf{2.6{\pm0.1}}$ & $\mathbf{2.6{\pm0.1}}$\\
Mudline shear                & $6.8{\pm0.3}$ & $6.8{\pm0.2}$ & $6.4{\pm0.1}$ & $6.1{\pm0.1}$ & $\mathbf{6.0{\pm0.1}}$ & $6.2{\pm0.1}$\\
Mudline moment               & $3.3{\pm0.3}$ & $3.3{\pm0.1}$ & $\mathbf{3.1{\pm0.1}}$ & $3.2{\pm0.0}$ & $\mathbf{3.1{\pm0.2}}$ & $3.2{\pm0.1}$\\
Blade-root edgewise          & $5.3{\pm0.9}$ & $5.8{\pm0.3}$ & $5.6{\pm0.1}$ & $4.9{\pm0.0}$ & $\mathbf{4.5{\pm0.3}}$ & $4.7{\pm0.4}$\\
Blade-root flapwise          & $2.5{\pm0.2}$ & $2.5{\pm0.1}$ & $\mathbf{2.3{\pm0.1}}$ & $2.4{\pm0.1}$ & $\mathbf{2.3{\pm0.2}}$ & $2.4{\pm0.2}$\\
\multicolumn{4}{@{}l|}{\emph{Side--side channels}} & & &\\
Tower-base moment            & $61.7{\pm40.7}$ & $62.2{\pm11.3}$ & $10.3{\pm1.8}$ & $7.5{\pm0.4}$ & $8.1{\pm1.1}$ & $\mathbf{6.7{\pm0.4}}$\\
Mudline shear                & $12.2{\pm2.7}$ & $\mathbf{11.3{\pm1.5}}$ & $11.6{\pm0.4}$ & $12.7{\pm0.4}$ & $12.2{\pm0.6}$ & $12.5{\pm0.4}$\\
Mudline moment               & $44.5{\pm32.2}$ & $44.3{\pm9.0}$ & $8.8{\pm0.7}$ & $8.4{\pm0.3}$ & $\mathbf{8.0{\pm0.1}}$ & $8.2{\pm0.2}$\\
\cmidrule(l){1-7}
Mean (8 channels)            & $17.4{\pm9.6}$ & $17.4{\pm2.8}$ & $6.3{\pm0.2}$ & $6.0{\pm0.1}$ & $5.9{\pm0.0}$ & $\mathbf{5.8{\pm0.1}}$\\
\bottomrule
\end{tabular}
\end{table}

\begin{table}[htbp]
\centering
\caption{\textbf{Inference time and training memory.} End-to-end rollout time per test record (mean $\pm$ s.d. over $20$ records) and peak GPU memory of one training step (batch $12$); one training seed per operator.}
\label{stab:cost}
\begin{tabular}{l cc}
\toprule
operator & end-to-end rollout [s] & training memory [GiB]\\
\midrule
DeepONet & $0.443\pm0.008$ & $2.21$\\
FNO & $0.412\pm0.001$ & $2.20$\\
DeepOT & $0.550\pm0.012$ & $9.28$\\
DeepOMamba & $0.480\pm0.007$ & $7.97$\\
DeepOMamba$_{64}$ & $0.538\pm0.002$ & $9.39$\\
DeepOMamba$_{565}$ & $0.498\pm0.001$ & $10.64$\\
\cmidrule(l){1-3}
DeepOJamba & $0.495\pm0.006$ & $8.11$\\
DeepOJamba$_{64}$ & $0.549\pm0.010$ & $9.17$\\
\bottomrule
\end{tabular}
\end{table}

\begin{table}[htbp]
\centering
\caption{\textbf{Settings of the operators.} In the convolution-width study, $d_{\mathrm{conv}}$ is $64$ or $565$, and the $565$-step models have hidden width $168$ (DeepOMamba) or $176$ (DeepOJamba).}
\label{stab:internals}
\renewcommand{\arraystretch}{1.15}
\begin{tabular}{p{0.38\linewidth}p{0.50\linewidth}}
\toprule
setting & value\\
\midrule
blocks & DeepOT $L=9$ attention; DeepOMamba $L=8$ Mamba; DeepOJamba \texttt{MMAMMAMM} ($L_{\mathrm{M}}=6$, $L_{\mathrm{A}}=2$)\\
hidden width $d$ & $192$\\
feed-forward & SwiGLU, $4d$\\
normalization & pre-RMSNorm\\
attention & full, $H=4$ heads, RoPE, query--key norm, FlashAttention\\
Mamba state size $d_{\mathrm{state}}$ & $16$\\
Mamba $d_{\mathrm{conv}}$ & $4$ (Mamba default)\\
Mamba expansion & $2$\\
$\Delta$ projection rank $\lceil d/16\rceil$ & $12$\\
$\Delta$ initialization & $\mathrm{softplus}(\mathbf{b}_\Delta)$ log-uniform in $[10^{-3},10^{-1}]$\\
$\mathbf{A}$ initialization & $-\mathrm{diag}(1,\dots,d_{\mathrm{state}})$ per channel (S4D-real)\\
\midrule
DeepONet baseline & branch: strided one-dimensional CNN over the per-step forcing and wind features, $16$-channel bottleneck pooled to $256$ time bins (preserving forcing timing), fused with the start state and case descriptors; trunk: Fourier-feature network of time with $128$ frequencies; $P=256$ basis functions; $6.08$\,M parameters\\
FNO baseline & four Fourier layers of width $76$ retaining $k_{\max}=128$ modes (up to $3.125$~Hz over the $40.96$~s window); case descriptors concatenated to every step; $6.02$\,M parameters\\
\bottomrule
\end{tabular}
\end{table}

\begin{table}[htbp]
\centering
\caption{\textbf{Conditioning placement on the ten-story shear building.} Test \relltwo{}, averaged over story displacement, velocity, absolute acceleration, and restoring force, for bilinear and Bouc--Wen story laws (one training seed). Bold: lowest per row; $-$: not applicable or not run}
\label{stab:cond}
\begin{tabular}{l l ccccc}
\toprule
law & operator & block-level & sublayer-level & mixed & concat & late\\
\midrule
bilinear & DeepOT & $0.0617$ & $\mathbf{0.0535}$ & $-$ & $0.0550$ & $0.1081$\\
 & DeepOMamba & $\mathbf{0.0354}$ & $0.0390$ & $-$ & $0.0405$ & $0.0941$\\
\cmidrule(l){2-7}
 & DeepOJamba & $0.0398$ & $-$ & $\mathbf{0.0387}$ & $0.0467$ & $0.0992$\\
\midrule
Bouc--Wen & DeepOT & $0.0392$ & $\mathbf{0.0354}$ & $-$ & $0.0364$ & $0.0617$\\
 & DeepOMamba & $\mathbf{0.0228}$ & $0.0251$ & $-$ & $0.0258$ & $0.0498$\\
\cmidrule(l){2-7}
 & DeepOJamba & $\mathbf{0.0248}$ & $-$ & $0.0258$ & $0.0308$ & $0.0519$\\
\bottomrule
\end{tabular}
\end{table}

\clearpage
\bibliographystyle{sn-nature}
\bibliography{sn-bibliography}